\def\year{2019}\relax
\documentclass[letterpaper]{article}
\usepackage{aaai19}
\usepackage{times}
\usepackage{helvet}
\usepackage{courier}
\usepackage{mathtools}
\usepackage{amsthm}

\usepackage{float}
\usepackage[colorlinks,linkcolor=red,anchorcolor=red,citecolor=red]{hyperref}
\usepackage[]{graphicx}
\usepackage{subfigure}
\usepackage{multicol}
\usepackage{paralist}
\usepackage{bm}
\usepackage{amsfonts}
\usepackage{url}
\usepackage{multirow}
\usepackage{cases}
\theoremstyle{definition}

\usepackage{booktabs}
\usepackage{threeparttable}
\usepackage{color}

\makeatletter
\newif\if@restonecol
\makeatother

\usepackage[linesnumbered,ruled,vlined]{algorithm2e}%[ruled,vlined]{
\usepackage{algpseudocode}
\usepackage{amsmath}
\begin{document}
\title{Interactive Binary Image Segmentation with Edge Preservation}
%\author{Paper ID \#4168}
\author{Jianfeng Zhang\\
	Farsee2 Technology and School of Mathematics and Statistics, Wuhan University, China\\
	\AND
	Liezhuo Zhang \\ 
	School of Computer Science and Technology, Huazhong  University of Science and Technology\\
	\AND
	Yuankai Teng \and  Xiaoping Zhang \\ 
	School of Mathematics and Statistics, Wuhan University
	\AND
	Song Wang \\ 
	Department of Computer Science and Engineering, University of South Carolina, USA and Farsee2 Technology, China\\
	\AND
	Lili Ju \\ 
	Department of Mathematics, University of South Carolina, USA and Farsee2 Technology, China
}

\maketitle
\begin{abstract}
	Binary image segmentation plays an important role in computer vision and has been widely used in many applications such as image and video editing, object extraction, and photo composition. In this paper,  we propose a novel interactive binary  image segmentation method based on the Markov Random Field (MRF) framework and the fast bilateral solver (FBS) technique. Specifically, we employ the geodesic distance component to build the unary term. To ensure both computation efficiency and effective responsiveness for interactive segmentation, superpixels are used in computing geodesic distances instead of pixels. Furthermore, we take a bilateral affinity approach for the pairwise term in order to preserve edge information and denoise. Through the alternating direction strategy, the MRF energy minimization problem is divided into two subproblems, which then can be easily solved by steepest gradient descent (SGD) and FBS respectively. Experimental results on the VGG interactive image segmentation dataset show that the proposed algorithm outperforms several state-of-the-art ones, and in particular, it can achieve satisfactory edge-smooth segmentation results even when the foreground and background color appearances are quite indistinctive.
\end{abstract}

\section{Introduction}
\noindent Binary image segmentation,  the process of partitioning a digital image into foreground and background regions, is one of the most fundamental problems in computer vision, which has been widely used in many  applications, such as image and video editing, object extraction and recognition, photo composition, medical image analysis, and so on. Automatic image segmentation sometimes could not  produce satisfactory results  due to the fact that the foreground object is really ambiguous if any prior knowledge or high-level understanding of the content is absent. Therefore, most of existing binary image segmentation algorithms allow  user-provided interactions in order to obtain information about the object of interest, such as scribbles \cite{bai2007,boykov2001,grady2006,protiere2007,wang2007} or  bounding boxes \cite{rother2007,lempitsky2009,cheng2015}. Through these interactions from users,   high-level semantic knowledge could be obtained, further leading to desired binary segmentation results.

\begin{figure}[!htbp]
	\hspace*{-.2cm}
	\centering
	\subfigure[]{%
		\includegraphics[width=0.211\linewidth]{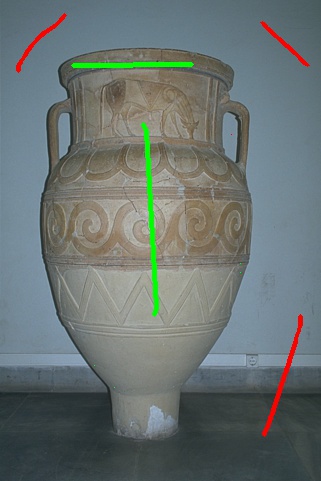}
	}
	\subfigure[]{%
		\includegraphics[width=0.37\linewidth]{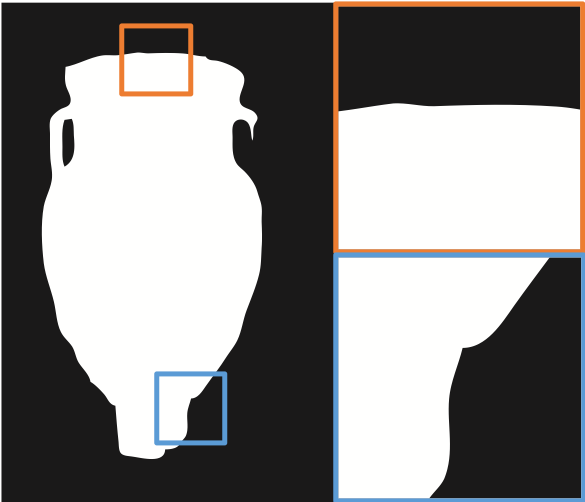}
	}
	\subfigure[]{%
		\includegraphics[width=0.37\linewidth]{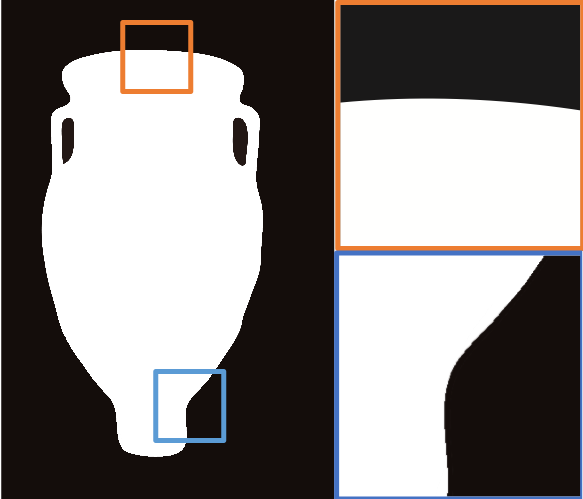}
	}
	\caption{(a) A RGB image with scribbles where green ones denote the foreground and red ones the background; (b) segmentation by graph cut; (c) segmentation  by the proposed method.}
	\label{fig:figure1}
\end{figure}

In general, there are two basic requirements for a good interactive image segmentation algorithm: 1) given a certain user input, the algorithm should produce accurate segmentation results that reflect the user intent; 2) user interaction should not be too complicated. To these ends, we adopt ``clicks'' or ``scribbles'' on the desired foreground and background regions (see Figure \ref{fig:figure1}) as effective user interactions, which can be easily controlled in most cases.

The emergence of many interactive segmentation algorithms has been witnessed in recent years. A popular framework is the total variation model \cite{unger2008,kwon2013,shi2016}. This model consists of an unary term that uses foreground and background colors inferred from the respective seed pixels, and a total variation term to localize edge. While the total variation term can explicitly refine object boundaries, it does not use any color information as guidance,  leading to unsatisfactory results in many cases. Another popular method is the graph cut method first introduced by \cite{boykov2001}, with numerous  variations (e.g. geodesic graph cut \cite{price2010}). However, graph cut method may suffer from the shrink bias toward shorter paths  \cite{price2010} because its pairwise term includes a summation over the boundary of the segmented regions. Geodesic graph cut seeks to solve this problem by using the geodesic distance instead of Euclidean distance as unary term, but  computing geodesic distance over the whole image is time-consuming. In addition to the forementioned approaches, there are also some image segmentation methods based on superpixels instead of pixels because superpixels can help reduce the computation complexity and thus accelerate the algorithms \cite{wang2011,schick2012,papazoglou2013,rantalankila2014,wang2015,khoreva2017}. Although the use of superpixels can release the computational cost, it may produce inaccurate boundaries on the other hand. Therefore, image segmentation methods based on superpixels often need to apply some post-processing to refine the boundaries \cite{schick2012,feng2016}.

In this work, we aim to design an interactive binary image segmentation method which can achieve high quality segmentation results. To these ends, we will formulate our task as a binary labeling problem via Markov Random Field (MRF) with the unary and pairwise terms to model image segmentation.  
It has been shown that fast bilateral solver (FBS) \cite{barron2016}, a novel algorithm for edge-aware smoothing developed  recently, can help to denoise and preserve object boundaries efficiently. Therefore, we further take a bilateral affinity term as the pairwise term in the MRF framework in order to obtain bilateral-smooth results. Through the alternating direction strategy, we are able to apply steepest gradient descent (SGD) and FBS together to effectively solve the target energy minimization problem. To sum up, our method  includes the following major components and advantages:

\begin{itemize}
	
	\item Geodesic distance is employed to compute unary term to separate regions with similar color appearance that belong to different labels. To ensure algorithm efficiency, we compute geodesic distance in the superpixel level to dramatically release the computational burden. 
	
	\item The bilateral affinity is used as the pairwise term, which produces segmentation results with good boundary sensitivity regardless of image complexity.
	
	\item The overall optimization problem is split into two subproblems through alternating direction approach, which can be effectively solved via SGD and FBS, respectively. 
	
\end{itemize}

\section{Related Work}
\subsection{Interactive image segmentation}
\noindent A large body of work has been proposed for interactive image segmentation on color images. Among them, the graph cut \cite{boykov2001} approach has been a very popular one, in which an image is represented as a graph and  a globally optimal segmentation based on the MRF energy minimization is found with balanced region and boundary information. However, graph cut is often limited because it only relies on color information and thus can fail in cases where the foreground and background color distributions are overlap or very complicated. In addition, graph cut may suffer from the aforementioned shrink bias problem \cite{price2010}. 

Another effective framework is to use the geodesic information. A geodesic image segmentation algorithm was proposed in \cite{criminisi2008}, in which  image segmentation is viewed as an approximate energy minimization problem in a conditional random field, and the segmentation task is then finished by expanding outward from the seeds to selectively fill the desired region. Geodesic  information is useful for selecting objects with complex boundaries such as those with long and thin parts. However, since it only works from the interior of the selected object outwards and does not explicitly consider the object boundary, it may suffer from a bias that favors shorter paths back to the seeds. Another interactive segmentation method based on geodesic information is the geodesic graph-cut introduced in \cite{price2010}. It combines geodesic distance-based region information with edge information in the graph-cut optimization framework. It also introduces a spatially-varying weighting scheme based on the local confidence of the geodesic component, which is used to adjust the relative weighting between the binary and pairwise terms. Geodesic distance component effectively helps  avoid the tendency for geodesic segmentation to degenerate to Euclidean distance maps when the foreground/background colors are indistinct.

There are also some interactive image segmentation methods performing on the superpixels level instead of the pixel level since superpixels can groups pixels into perceptually meaningful regions and  greatly reduce the computation complexity. \cite{schick2012} ameliorated foreground segmentation through 
post-processing based on superpixels. This method first converts the pixel-based segmentation into a probabilistic superpixel representation and then use MRF to refine segmentation. In \cite{feng2016} superpixels are used  instead of pixels as graph nodes in the modified graph cut algorithm in order to ensure good responsiveness and efficiency of interactive segmentations. 

\subsection{Regularization techniques}
Lot of researches related to image segmentation impose the regularization requirement of the results by  adding a  pairwise term into the models \cite{Chartrand2008,lam2010,zhu2013,zhang2015}.  The Rudin-Osher-Fatemi and total variation regularizations were used in \cite{Chartrand2008}, which can preserve and smooth boundary edges.  Euler's elastica was applied in \cite{zhu2013} as regularization of the activate contour segmentation model. Although the modified activate contour model is able to preserve local boundaries as well as capture fine elongated structures of objects, it still encounter the problem of omitting relatively small objects. In \cite{zhang2015}  an image segmentation method was proposed, which adopts a sparse and low-rank based nonconvex regularization. This model can capture the global structure of the whole data, often leading to better segmentation results than the total variation based model. All aforementioned regularization methods are only based on the pixel label similarity without taking account of image color information. 
Bilateral affinity  is also a  regularization that can be added into the image segmentation models. It combines color and spatial information to locate the object boundary and denoise in order to produce smooth segmentation results. Fast bilateral solver \cite{barron2016}, a novel algorithm that can  very efficiently compute the bilateral affinity term, has been applied in various computer vision tasks, such as stereo, depth super-resolution and colorization, to produce edge-aware smoothing results. We will use the  bilateral affinity regularization in the proposed algorithm. 

\section{The proposed method}

The proposed method  is based on solving the minimization problem of the following energy functional within the  MRF framework:
\begin{equation}\label{energy}
\min_{\boldsymbol{u}} \mathcal E(\boldsymbol{u}) = \sum_{i \in \Omega} \mathcal R(u_i) + \lambda \sum_{(i,j)\in \mathcal N } W_{ij} (u_i-u_j)^2, 
\end{equation}
where $\Omega$ is the set of pixels of the given image, $\mathcal N$ is the set of pairs of neighboring pixels,  $\lambda$ is a regularizing  parameter, and $\boldsymbol{u}=(u_i)$ is a binary vector of labels defined by
$$
u_i=\left\{
\begin{array}{ll}
1, & \mbox{if} \ \  i \in \Omega_{\mathcal F}, \\[.05in]
0, & \mbox{if} \ \  i \in \Omega_{\mathcal B}, 
\end{array}
\right.
$$
with $\Omega_{\mathcal F}$  and $\Omega_{\mathcal B}$ denoting the set of  pixels  of the foreground  and background of the image, respectively. The first term and the second term in the righthand of \eqref{energy} are often referred as  the ``unary term''  and the  ``pairwise term'', respectively. In the binary segmentation task, the unary term represents the cost of assign label $u_i$ to pixel $i$, and is often formulated as follows: 
\begin{equation}
\mathcal R(u_i) = f_i^{(1)}u_i + f_i^{(2)}(1-u_i).
\end{equation}
How to choose $f_i^{(l)}, l=1,2$ is a core issue of the binary segmentation problem, and  two commonly-used forms can be presented as:
\begin{itemize}
	\item from the ``Gaussian color model'' for each region --
	\begin{equation}\label{dataterm1}
	f_i^{(l)} = \frac{(I_i-\mu_l)^2}{2\sigma_l^2}+\log \sigma_l, \ \ l=1,2,
	\end{equation}
	\item  from the ``general color distribution''  --
	\begin{equation}\label{dataterm2}
	f_i^{(l)} = -\log p_l(I_i), \ \ l=1,2,
	\end{equation}
\end{itemize}
where $I_i$ denotes the image density at the pixel $i$, $\mu_l$ and  $\sigma_l$ are the mean and variance of the forground/background seed sets, respectively, and $p_l(I_i)$ is the probability of the $i$-th pixel belonging to the forground/background. The pairwise term represents the cost for assigning a pixel pair $u_i$ and $u_j$, and is used to introduce some additional smooth constraints in the segmentation tasks, such as denoising,  edge-preserving, etc.

% \begin{rem}
% Graph cut segmentation \cite{boykov2001}, where $f_i^{(k)}$ is chosen by \eqref{dataterm2} and 
% $$
% W_{i,j} \propto \exp\left(\frac{-\|I_i-I_j\|^2}{2\sigma^2}\right).
% $$ 
% \end{rem}

In our method, the unary term is computed by utilizing the geodesic distance, accompanied by the superpixel-based acceleration, which aims at distinguishing the similar colors in the foreground and background regions. The pairwise term is generated by introducing the bilateral affinity as a regularizer, which is used to ensure the  edge-preservation. Finally, to efficiently solve the minimization of  \eqref{energy}, we adopt the alternating direction strategy to split the minimization problem of \eqref{energy} into two subproblems, and then iteratively solve them by SGD and FBS until  the solution converges. 

Next we present a detailed description of the proposed method and its implementation techniques.

\subsection{Unary term}
In the binary segmentation task,  users need to provide some cues to distinguish the foreground and background regions. Based on these cues, $f_i$ will be then determined by some ways. Although \eqref{dataterm1} and \eqref{dataterm2} are widely used in many interactive segmentation methods, their drawbacks are also obvious. For example, it could fail to separate the color-similar regions located at the foreground and background boundaries. To overcome this difficulty, we adopt the geodesic distance at the superpixel level to compute $f_i^{(l)}$. 

% The most popular approach to seeded segmentation may be the graph-cut method, and it have been widely used in computer vision. However, since it uses Euclidean distance to compute the unary term, it seems difficult to distinguish two objects with similar colors. In addition, graph-cut may suffer from shrink bias problem and thus leads to undesired segmentation results. To overcome this drawback, we take geodesic distance as an alternative to compute the unary term.

Let $\widehat{\Omega}_l$ be  the set of seeds with user annotated labels $l\in\{\mathcal F, \mathcal B\}$ where $\mathcal F$
and $\mathcal B$ stand for the foreground and the background respectively.
The geodesic distance from a pixel $i$ to  $\widehat{\Omega}_l$ is then defined by 
\begin{equation}
d(i, \widehat \Omega_l) = \min_{j\in \widehat \Omega_l} d(i,j), 
\end{equation}
where  $d(i,j)$ denotes the geodesic distance between the two pixels $i$ and $j$, which can be computed by the famous Dijkstra algorithm. 
However, the computation of geodesic distances at the pixel level is extremely time-consuming, and hence restrict its application in practice, especially when the image is very large. To reduce such  computational burden, we use the superpixels to approximate and accelerate computation of geodesic distances.

A superpixel can be defined as a group of pixels which have similar characteristics and it is generally a color-based segmentation. Superpixels have become the fundamental units in many imporatnt computer vision tasks. Our idea is described by the following steps: 1) generate a set of superpixels of the image, $\{S_k\}_{k=1}^K$, with centers $\{O_k\}_{k=1}^K$ by an existing method, such as SLIC \cite{achanta2010}, HEWCVT \cite{zhou2015} and so on; 2) for $k=1,\cdots, K$, compute $d(O_k , \widehat \Omega_l) $, the geodesic distance from the center of the superpixel $S_k$ to the foreground-background seed set;  3) obtain $f_i^{(l)} = d(O_k , \widehat \Omega_l)$ if $i \in S_k$ in the unary term.  

\subsection{Pairwise term}
The generation of superpixels is fast, and relieve the computational burden of  geodesic distances dramatically. However, there also exist some drawbacks in practice. For example, superpixels often cross the edges between two color-similar objects, which directly affect the final foreground/background segmentation results. To solve this drawback, we choose  $\boldsymbol{W}$ in \eqref{energy} as the bilateral affinity matrix. Each element of the bilateral affinity matrix $W_{ij}$ reflects the affinity between pixels $i$ and $j$ in the reference image in the YUV colorspace:  \begin{equation}\label{BF}
\begin{aligned}
W_{ij} & =\exp\left( -\frac{\|\boldsymbol{p}_i^{x,y}-\boldsymbol{p}_j^{x,y}\|^2}{2\sigma_{xy}^2}
-\frac{(p_i^l-p_j^l)^2}{2\sigma_l^2} \right. \\
& \hspace{1.5in} \left.-\frac{\|\boldsymbol{p}_i^{u,v}-\boldsymbol{p}_j^{u,v}\|^2}{2\sigma_{uv}^2}\right),
\end{aligned}
\end{equation}
where  $\boldsymbol{p}_i$ is a pixel in the reference image $\boldsymbol{p}$ with the spatial position $\boldsymbol{p}_i^{x,y}=(p_i^x,p_i^y)$ and the color $\boldsymbol{p}_i^{l,u,v}=(p_i^l, p_i^u,p_i^v)$ (for clearness, we denote $\boldsymbol{p}_i^{u,v}=(p_i^u,p_i^v)$),  and  the parameters $\sigma_{xy}, \sigma_l$ and $\sigma_{uv}$  {control the extent of the spatial, luma, and chroma support of the filter, respectively}.   The choice of \eqref{BF} leads to edge-preserving to some extent. 

%
%\subsubsection{Total variation}
%
%Total variation 

\subsubsection{Fast bilateral solver}

%Direct computation of $\boldsymbol{W}$  We adopt t

Fast bilateral solver,  proposed in \cite{barron2016},  is a novel algorithm for edge-aware smoothing that combines the flexibility and speed of simple filtering approach with the accuracy of the domain-specific optimization algorithm. 
FBS attempts to minimize the following functional 
\begin{equation}\label{optim1}
\min_{\boldsymbol{v}} \sum_{i} c_i(v_i-t_i)^2 + \frac\lambda2\sum_{i,j}  W_{ij} (v_i-v_j)^2, 
\end{equation}
where $\boldsymbol{t} = (t_i)$ is an input target vectorized image, $\boldsymbol{c}=(c_i)$ is a confidence vectorized image,  $\lambda$ is the pairwise term multiplier, and $W_{ij}$ is defined by \eqref{BF}, which is the bilateral filter weight for the pixel pairs $(i,j)$ given a reference RGB image $\boldsymbol{p}$. 

Direct solution of \eqref{optim1} is generally computationally expensive, especially when the image has high resolution. 
%Instead, FBS treat it in the associated so-called bilateral space. 
%%  the smoothness term is built around an affinity matrix $\widehat \boldsymbol{W}$, which is a bistochastized version of a bilateral affinity matrix  $\boldsymbol{W}$ defined in \eqref{BF}. 
There are techniques for speeding up bilateral filtering. Two of them, the permutohedral lattice \cite{adams2010} and the bilateral grid \cite{bai2007} express bilateral filtering as a splat/blur/slice procedure: 1) pixel values are {``splatted''} into a small set of vertices in a grid; 2) those values are {``blurred''} in bilateral space; 3) the filtered pixel values are produced via a {``slice''} (an interpolation) of the blurred vertex values.
These approaches correspond to a compact and efficient factorization of $\boldsymbol{W}$:
\begin{equation}\label{W_fact}
\boldsymbol{W} \approx \widetilde{\boldsymbol{W}} = \boldsymbol{S}^T  \boldsymbol{B} \boldsymbol{S},
\end{equation}
where the multiplication by $\boldsymbol{S}$ is the ``splat'', the multiplication by $ \boldsymbol{B}$ is the ``blur'', and the multiplication by $\boldsymbol{S}^T$ is the ``slice''.
% \textcolor{blue}{!!!what are $\boldsymbol{S}$ and $\bar \boldsymbol{B}$!!!}
The factorization \eqref{W_fact} allow for the optimization problem \eqref{optim1} to be ``splatted'' and solved in the bilateral space. {A bistochasticized version of $\widetilde{\boldsymbol{W}}$ can be obtained  on  the 
	``simplified'' bilateral grid \cite{barron2015}: 
	\begin{equation}\label{W_hat}
	\widehat{\boldsymbol{W}} = \boldsymbol{S}^T \boldsymbol{D}_{\boldsymbol{m}}^{-1}\boldsymbol{D}_{\boldsymbol{n}} \boldsymbol{B} \boldsymbol{D}_{\boldsymbol{n}}\boldsymbol{D}_{\boldsymbol{m}}^{-1}\boldsymbol{S}, 
	\end{equation}
	where 
	$\boldsymbol{D}_{\boldsymbol{n}}$ and $\boldsymbol{D}_{\boldsymbol{m}}$ are two bistochastization matrices
	and $ \boldsymbol{S}$ satisfies
	$$ 
	\boldsymbol{S}\boldsymbol{S}^T = \boldsymbol{D}_{\boldsymbol{m}}.
	$$}
%\textcolor{blue}{!!!what are $\boldsymbol{W}$ and $\boldsymbol{B}$!!!}
Replacing the weight matrix in \eqref{optim1} by \eqref{W_hat} leads to the following approximate optimization problem:
\begin{equation}\label{optim2}
\min_{\boldsymbol{v}} \sum_{i} c_i(v_i-t_i)^2 + \frac\lambda2\sum_{i,j}  \widehat W_{ij} (v_i-v_j)^2, 
\end{equation}
%The data term and  the smoothness term in \eqref{optim2}  can be reformulated as
%$$
%  %\frac\lambda2\sum_{i,j} \widehat W_{ij} (z_i-z_j)^2 = 
%  \lambda\boldsymbol{v}^T (\boldsymbol{I} -  \widehat\boldsymbol{W}) \boldsymbol{v}, 
%$$
%and
%$$
%  %\sum_{i} c_i(z_i-t_i)^2 = 
%  \boldsymbol{v}^T\mathrm{diag}(\boldsymbol{c})\boldsymbol{v}-2(\boldsymbol{c}\circ\boldsymbol{t})^T\boldsymbol{v}+(\boldsymbol{c}\circ\boldsymbol{t})^T\boldsymbol{t},
%$$  
%respectively. As a consequence, \eqref{optim2} can be reformulated aswhos
whose matrix form is 
\begin{equation}\label{optim3}
\begin{aligned}
&\min_{\boldsymbol{v}} ~~  \boldsymbol{v}^T\left(\lambda\left(\boldsymbol{I}-\widehat{\boldsymbol{W}}\right)+\mathrm{diag}(\boldsymbol{c})\right)\boldsymbol{v} \\ 
& \hspace{1in} - 2(\boldsymbol{c}\circ\boldsymbol{t})^T\boldsymbol{v}+(\boldsymbol{c}\circ\boldsymbol{t})^T\boldsymbol{t}.
\end{aligned}
\end{equation}
By introducing the transformation $\boldsymbol{y}=\boldsymbol{S}\boldsymbol{v}$, the optimization \eqref{optim2} in terms of pixels $\boldsymbol{v}$ can be described as an optimization in terms of bilateral-space vertices $\boldsymbol{y}$ 
%$$
%\min_{\boldsymbol{y}} ~~ \frac12\boldsymbol{y}^T\underbrace{}_{\boldsymbol{A}}\boldsymbo_{\boldsymbol{m}}l{y}-\underbrace{}_{\boldsymbol{b}^T}\boldsymbol{y}+\underbrace{\frac12(\boldsymbol{c}\circ\boldsymbol{t})^T\boldsymbol{t}}_{\boldsymbol{c}}, 
%$$
%i.e., 
$$
\min_{\boldsymbol{y}} ~~ \frac12\boldsymbol{y}^T\boldsymbol{A}\boldsymbol{y}-\boldsymbol{b}^T\boldsymbol{y}+c,
$$
where 
\begin{eqnarray*}
	\boldsymbol{A}&= &\left(\boldsymbol{D}_{\boldsymbol{m}}-\boldsymbol{D}_{\boldsymbol{n}} \boldsymbol{B} \boldsymbol{D}_{\boldsymbol{n}}\right)+ diag(\boldsymbol{S}\boldsymbol{c}),\\
	\boldsymbol{b} &= & \boldsymbol{S}(\boldsymbol{c}\circ\boldsymbol{t}),\\
	c &= &\frac12(\boldsymbol{c}\circ\boldsymbol{t})^T\boldsymbol{t},
\end{eqnarray*}
and $\circ$ is the Hadamard product. 
Note that \eqref{optim2} is a quadratic optimization problem and its solution is equivalent to solving the sparse linear system
$$
\boldsymbol{A}\boldsymbol{y}=\boldsymbol{b}.
$$
A pixel-space solution $\widehat{\boldsymbol{v}}$ of \eqref{optim3} then can be obtained by simply slicing $\boldsymbol{y}$:
$$
\widehat{\boldsymbol{v}} = \boldsymbol{S}^T\boldsymbol{y} = \boldsymbol{S}^T\boldsymbol{A}^{-1}\boldsymbol{b}.
$$

\subsection{Solution  by alternating direction}
The minimization of the energy functional \eqref{energy} is equivalent to
\begin{equation}\label{optim}
\begin{aligned}
& \min_{\boldsymbol{u}}  \sum_{i \in \Omega} \mathcal R_{i}(u_i) + \lambda \sum_{(i,j)\in \mathcal N } W_{ij} (v_i-v_j)^2 \\
& \hspace{0in} \mbox{s.t.} \;\; \boldsymbol{u}=\boldsymbol{v}.
\end{aligned}
\end{equation} 
The above optimization problem \eqref{optim} can be solved effectively by the alternating direction procedure iteratively:  
\begin{itemize}
	\item fixing $\boldsymbol{v}$, solve the minimization problem
	\begin{equation}\label{ad1}
	\min_{\boldsymbol{u}} ~~\sum_{i \in \Omega} \mathcal R_{i}(u_i) + \frac{\theta}{2} (u_i-v_i)^2;
	\end{equation}
	\item fixing $\boldsymbol{u}$, solve the minimization problem
	\begin{equation}\label{ad2}
	\min_{\boldsymbol{v}} ~~ \frac{\theta}{2} \sum_{i \in \Omega} (v_i-u_i)^2 + \frac\lambda2\sum_{(i,j)\in \mathcal N }  W_{ij} (v_i-v_j)^2, 
	\end{equation}
\end{itemize}
where $\theta$ is a regularizing parameter. 
The sub-problem \eqref{ad1} can be easily solved by SGD, and \eqref{ad2} is just the same as \eqref{optim1}, which can be solved efficiently by FBS. We present the
whole segmentation algorithm in the following Algorithm \ref{alg}.

\begin{algorithm}\label{alg}
	\SetAlgoLined
	\KwData{ A given image and some user-provided clicks or scribbles defining $ \widehat\Omega_l $ for $l\in\{\mathcal F, \mathcal B\}$}
	\KwResult{The binary segmentation image $\boldsymbol{u}$}
	Generate superpixels $\{S_k\}_{k=1}^K$ with centers $\{O_k\}_{k=1}^K$ by an existing method (such as SLIC)\;
	Compute all the geodesic distances $d(O_k , \widehat \Omega_l) $ by Dijkstra algorithm\;
	Set $f_i^{(l)} = d(O_k , \widehat \Omega_l)$ if $i \in S_k$ in the unary term $\mathcal R(u_i)$\;
	\While{not converged}{
		Use steepest gradient descent method to solve \eqref{ad1}\; 
		Use fast bilateral solver to solve \eqref{ad2}\;
		% Update the unary term by ...
	} 
	\caption{Interactive binary image segmentation with superpixel-geodesics and bilateral filter}
\end{algorithm}

\section{Experimental results}

In this section, we first introduce the testing dataset and then  compare the proposed method with several  state-of-the-art ones on the interactive binary  segmentation performance based on commonly-used evaluation criterions. In addition, we also give an ablation study for the proposed method. 

\subsection{Dataset}
In this study, we test the proposed method on the VGG interactive image segmentation dataset, provided by Visual Geometry Group, University of Oxford (\href{http://www.robots.ox.ac.uk/~vgg/data/iseg}{http://www.robots.ox.ac.uk/\~{}vgg/data/iseg}) \cite{Gulshan2010}. 
This dataset contains 151 images and the ground truth (GT) of segmentations. In addition, the dataset also provides a simple annotation for each image.
In detail, there are 49 images from GrabCut, 99 images from PASCAL VOC'09 and 3 images from the alpha-matting dataset. 
Images from the GrabCut dataset contain complex shapes but the foreground and background tend to have disjoint color distributions.
The VOC dataset on the other hand has simpler shapes (e.g., car, bus) but more complex appearances, where the color distributions of the foreground and background are overlap. 
The given annotations for images in the dataset are quite simple, which  sometimes do not offer sufficient semantic information for some complicated images  and correspondingly lead to undesired segmentation results for all compared methods as we find.  Thus we also add some extra clicks or scribbles for  these images, which produce a set of slightly modified  annotations, expecting to improve the segmentation results. 
%\textcolor{blue}{!!!Can we give a link for reader to download the new annotations?!!!}. 

% \subsection{Evaluating Criterions}
% We use some statistics, such as the \emph{IoU}, \emph{error rate} and \emph{$F_2$-score},  to evaluate the performance of a segmentation model. \emph{IoU} is defined by 
% \begin{equation}\label{miou}
% IoU = \frac {S_{Seg} \cap S_{GT}}{S_{Seg} \cup S_{GT}},
% \end{equation}
% where $S_{Seg}$ and $S_{GT}$ denote the area of segmentation result and the ground truth,  respectively. The numerator of \eqref{miou} can be also seen as the %overlap area of prediction and ground truth, and the denominator can be seen as the union area of that.

% The \emph{error rate} shows the proportion of error labeled pixels in all pixels of the image:

% $$error\ rate = \frac {the\ number\ of\ error\ labeled\ pixels} {the\ number\ of\ all\ pixels}$$

% \emph{$F_2$-score} is a weighted avarage to precision and recall, defined by
% $$
% F_2 = 5 \frac {precision \times recall}{(4 \times precision)+recall},
% $$
% where
% $$
% precision = \frac{TP}{TP+FP}, \ 
% recall = \frac{TP}{TP+FN}
% $$
% and $TP, FP$ and $FN$ denote true positive, false positive and false negative, respectively. 

\subsection*{Comparisons with other methods}

We use some statistics (quality measures), such as the \emph{IoU},  \emph{$F_2$-score}, \emph{error rate}, \emph{boundary precision}  and \emph{boundary recall}
%\textcolor{blue}{(see \href{https://www2.eecs.berkeley.edu/Research/Projects/CS/vision/bsds/}{https://www2.eecs.berkeley.edu/Research/Projects/CS/vision/bsds/}} 
to evaluate the performance of the proposed method (PM) and  compare it with some other well-known binary segmentation methods, such as geodesic graph cut (GEO), graph cut (GC), the total variation model using primal-dual method (TVPD), and the total variation model using alternative direction method (TVAD). The two total variation methods  use the same unary term as the proposed method, but the total variation as the pairwise term. 
For the proposed method, we use 1600 superpixels produced by SLIC to compute the geodesic distances (the effect of the number of superpixels on the segmentation performance will be investigated in the later subsection).  We also set $\lambda=100$ and $\theta=0.1$ in the propose method.

\begin{figure*}[!h]
	\begin{center}
		\hspace*{-.3cm}
		\begin{tabular}{ccccccccc}
			&\hspace{-0.25cm} OA & \hspace{-0.4cm}   MA & \hspace{-0.4cm}  OA & \hspace{-0.4cm}  MA&OA &\hspace{-0.4cm} \hspace{-0.4cm}   MA & \hspace{-0.4cm}  OA & \hspace{-0.4cm}  MA\\
			{\rotatebox[origin=t]{90}{\hspace{1cm}RGB}}&\hspace{-0.25cm}\includegraphics[width=.115\textwidth]{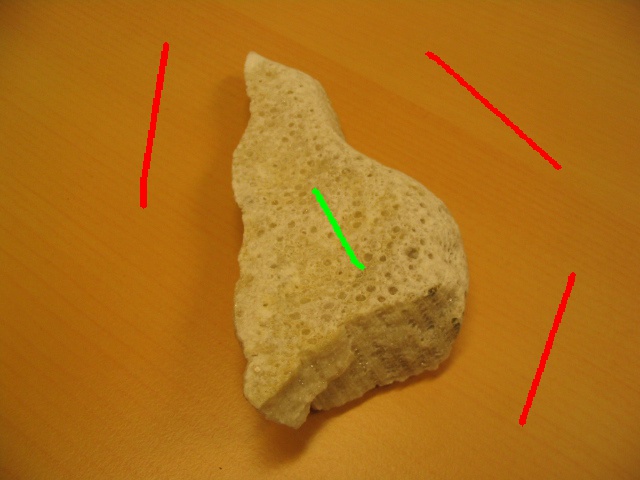} & \hspace{-0.4cm}
			\includegraphics[width=.115\textwidth]{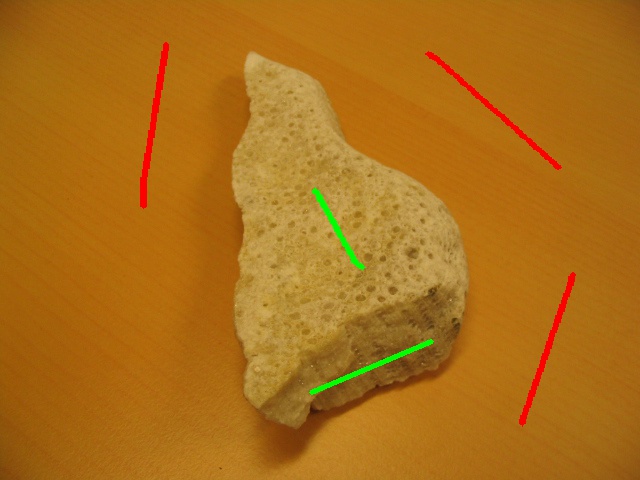} & \hspace{-0.3cm}
			\includegraphics[width=.115\textwidth]{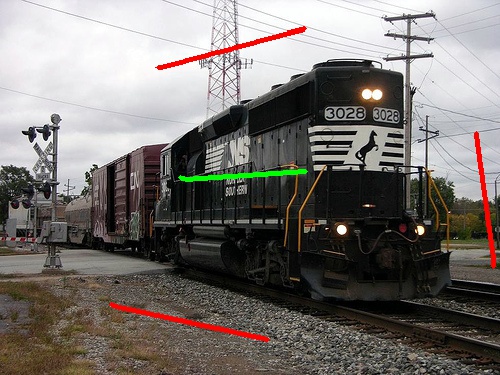} & \hspace{-0.4cm}
			\includegraphics[width=.115\textwidth]{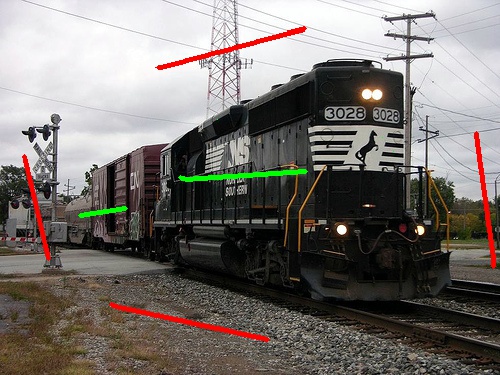}&\hspace{-0.3cm}
			\includegraphics[width=.13\textwidth]{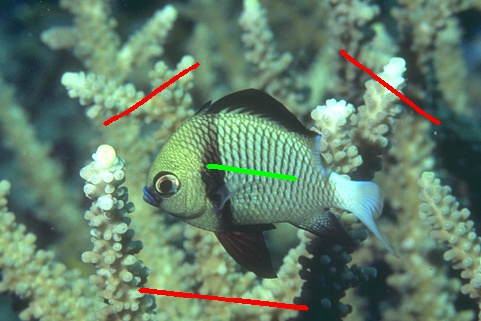} & \hspace{-0.4cm}
			\includegraphics[width=.13\textwidth]{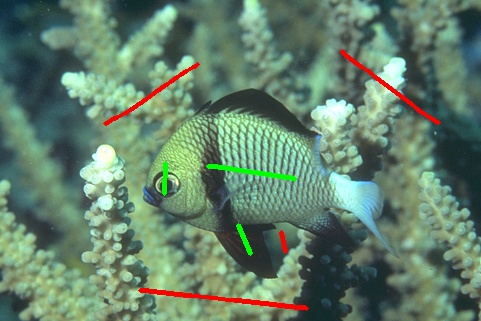} & \hspace{-0.3cm}
			\includegraphics[width=.1\textwidth]{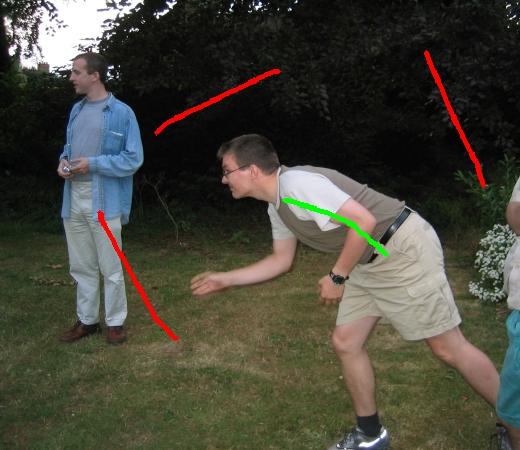} & \hspace{-0.4cm}
			\includegraphics[width=.1\textwidth]{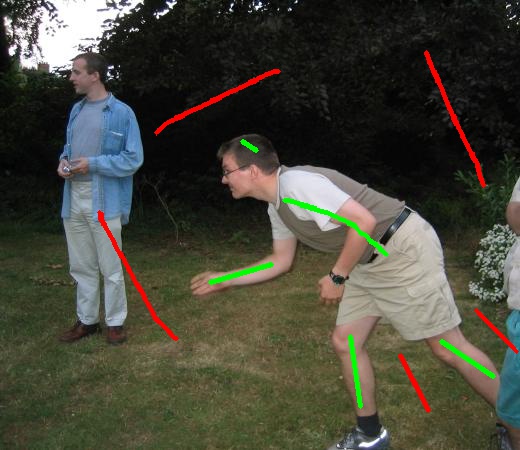}\\[-0.2in]
			{\rotatebox[origin=t]{90}{\hspace{1cm}GT}}&\hspace{-0.25cm}\includegraphics[width=.115\textwidth]{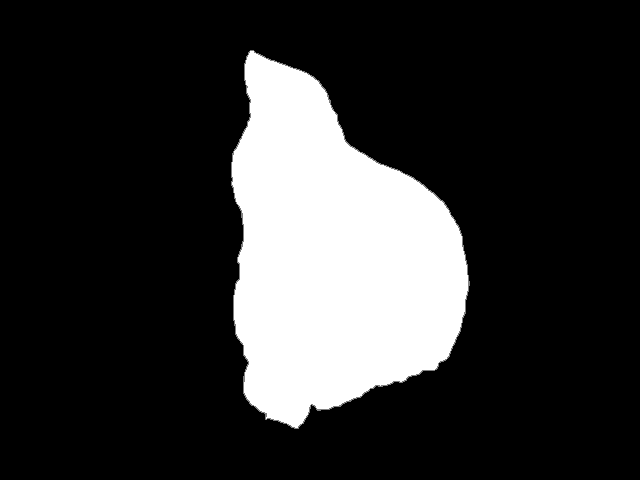} & \hspace{-0.4cm}
			\includegraphics[width=.115\textwidth]{images/fig2/stone_GT.png} &\hspace{-0.3cm} 
			\includegraphics[width=.115\textwidth]{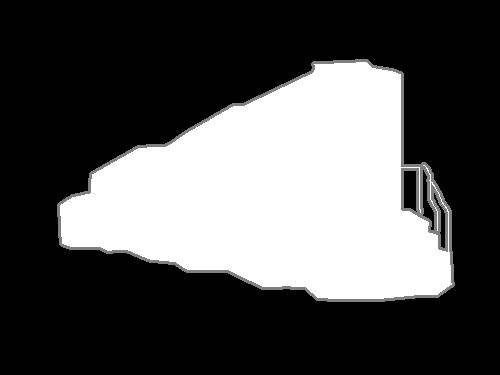} & \hspace{-0.4cm}
			\includegraphics[width=.115\textwidth]{images/fig2/train_GT.png}& \hspace{-0.3cm}
			\includegraphics[width=.13\textwidth]{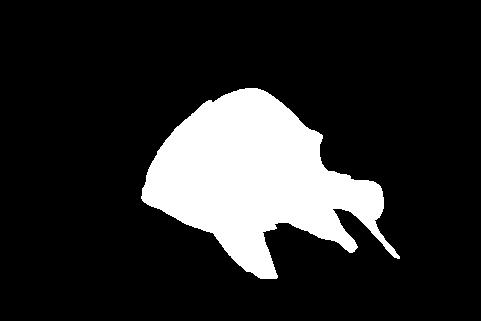} & \hspace{-0.4cm}
			\includegraphics[width=.13\textwidth]{images/fig2/209070_GT.png} &\hspace{-0.3cm} 
			\includegraphics[width=.1\textwidth]{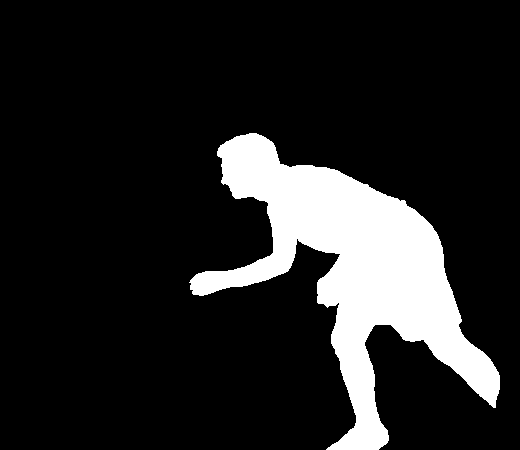} & \hspace{-0.4cm}
			\includegraphics[width=.1\textwidth]{images/fig2/bool_GT.png}\\[-0.15in]
			{\rotatebox[origin=t]{90}{\hspace{1cm}PM}}&\hspace{-0.25cm}\includegraphics[width=.115\textwidth]{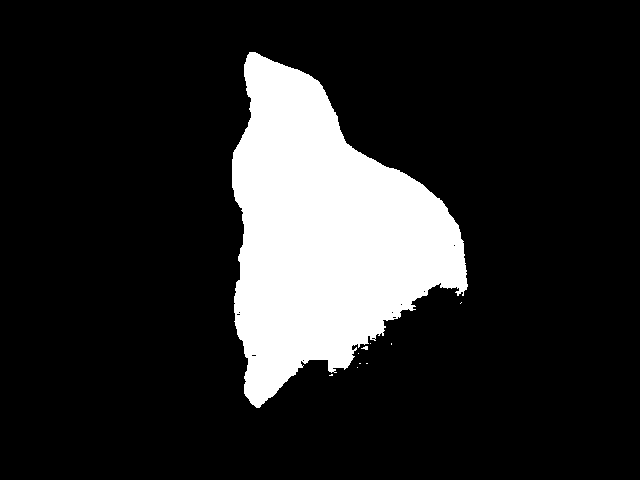} &\hspace{-0.4cm} 
			\includegraphics[width=.115\textwidth]{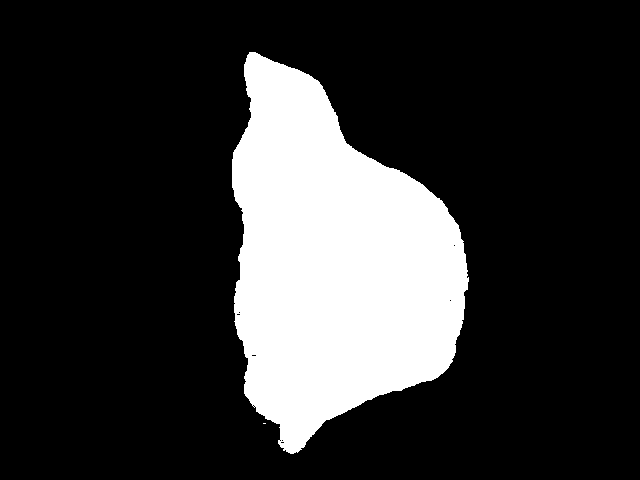} & \hspace{-0.3cm}
			\includegraphics[width=.115\textwidth]{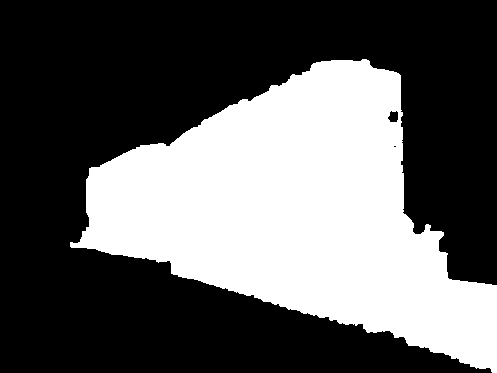} & \hspace{-0.4cm}
			\includegraphics[width=.115\textwidth]{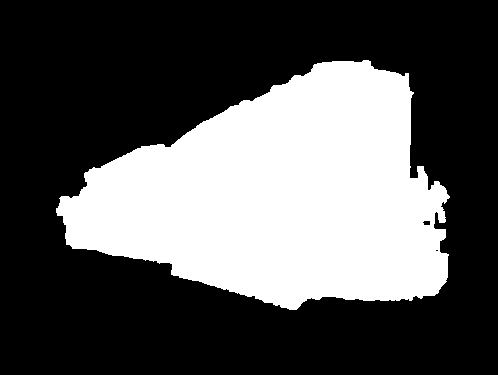}& \hspace{-0.3cm}
			\includegraphics[width=.13\textwidth]{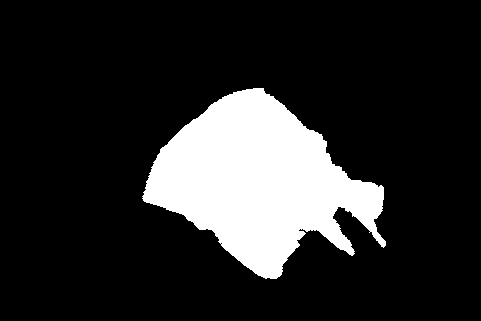} &\hspace{-0.4cm} 
			\includegraphics[width=.13\textwidth]{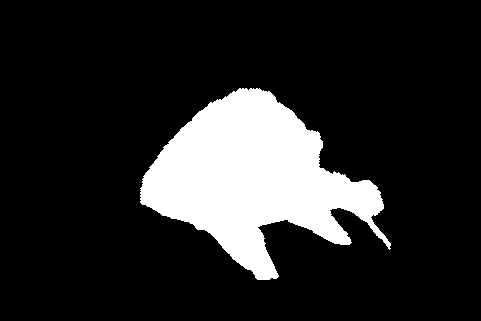} & \hspace{-0.3cm}
			\includegraphics[width=.1\textwidth]{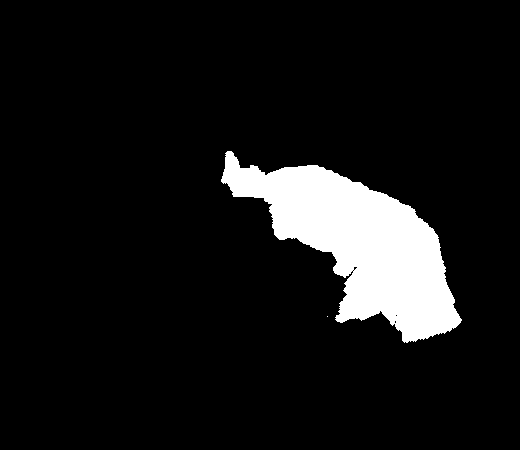} & \hspace{-0.4cm}
			\includegraphics[width=.1\textwidth]{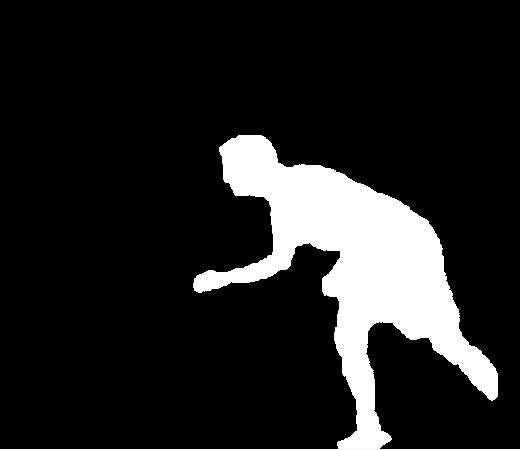}\\[-0.16in]
			{\rotatebox[origin=t]{90}{\hspace{1cm}GEO}}&\hspace{-0.25cm}\includegraphics[width=.115\textwidth]{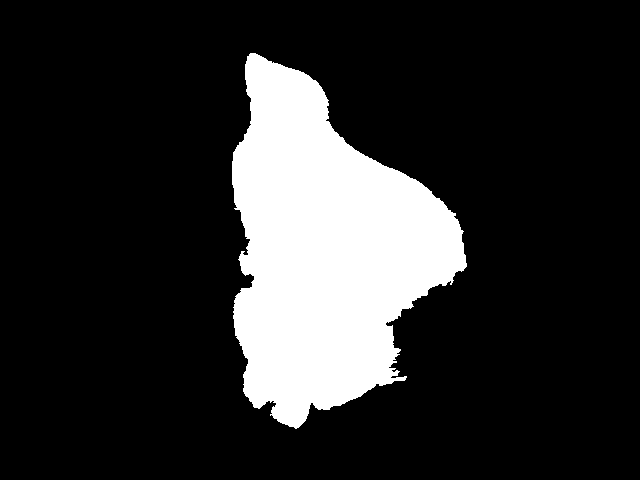} & \hspace{-0.4cm}
			\includegraphics[width=.115\textwidth]{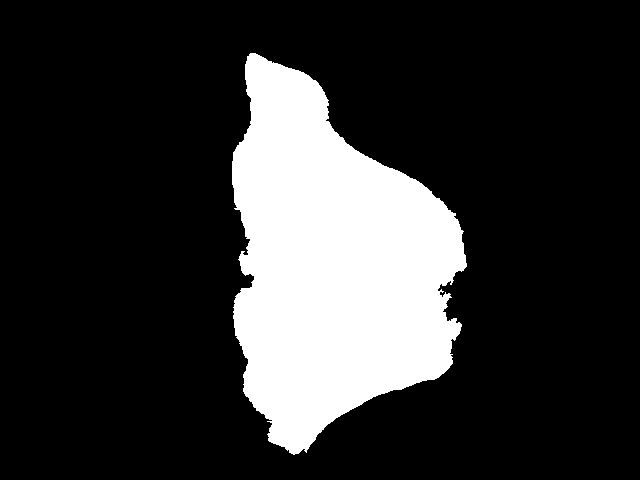} & \hspace{-0.3cm}
			\includegraphics[width=.115\textwidth]{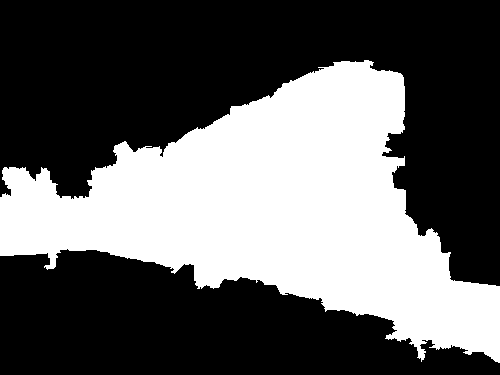} & \hspace{-0.4cm}
			\includegraphics[width=.115\textwidth]{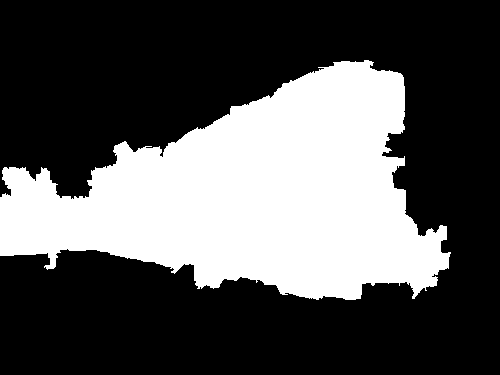}& \hspace{-0.3cm}
			\includegraphics[width=.13\textwidth]{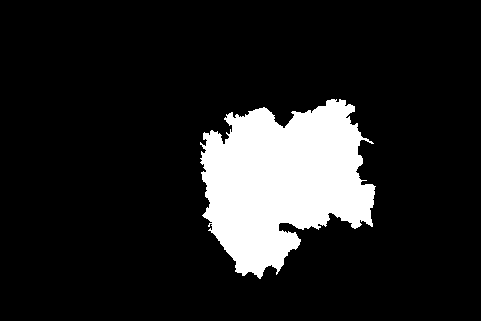} & \hspace{-0.4cm}
			\includegraphics[width=.13\textwidth]{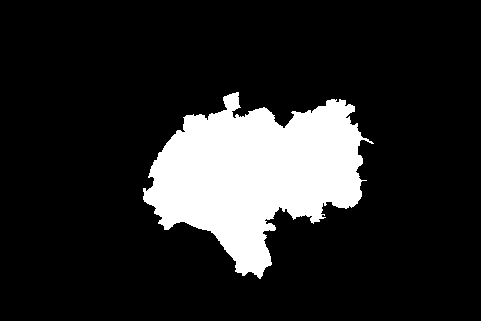} & \hspace{-0.3cm}
			\includegraphics[width=.1\textwidth]{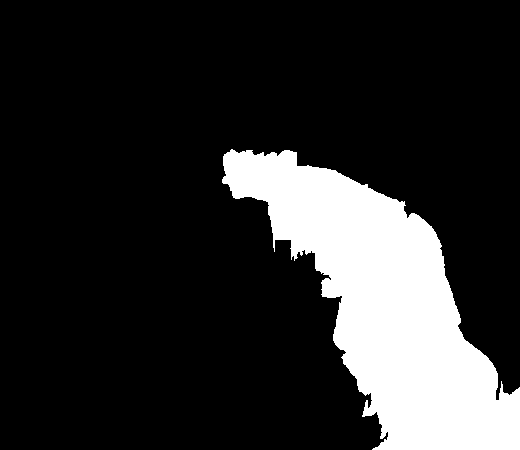} & \hspace{-0.4cm}
			\includegraphics[width=.1\textwidth]{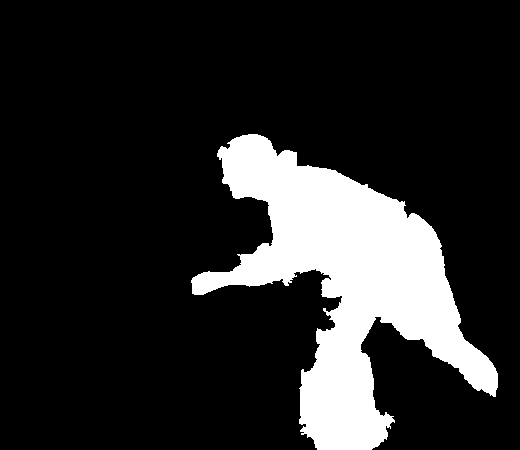}\\[-0.2in]
			{\rotatebox[origin=t]{90}{\hspace{1cm}GC}}&\hspace{-0.25cm}\includegraphics[width=.115\textwidth]{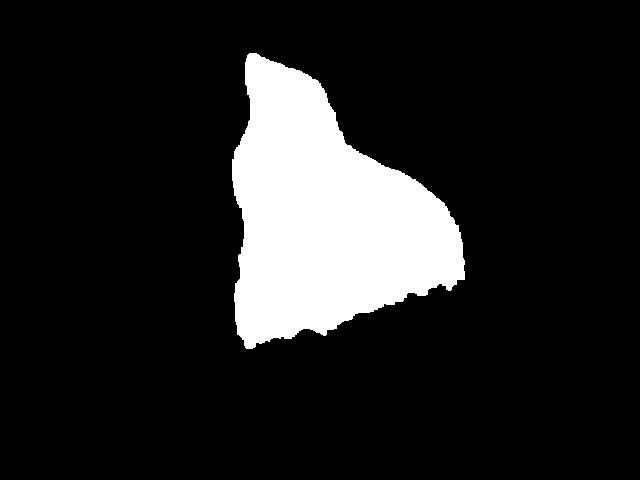} & \hspace{-0.4cm}
			\includegraphics[width=.115\textwidth]{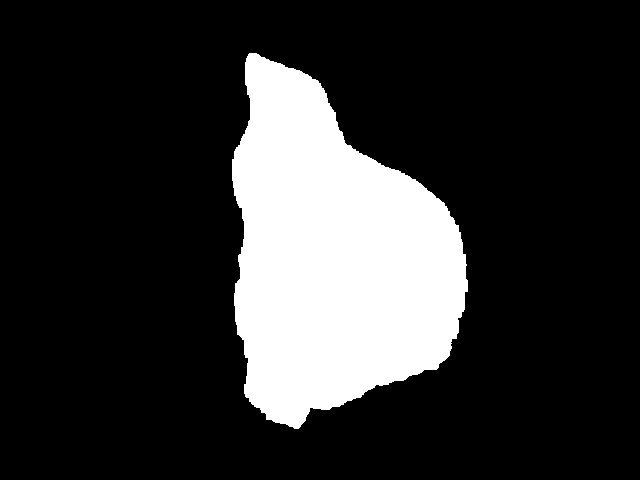} & \hspace{-0.3cm}
			\includegraphics[width=.115\textwidth]{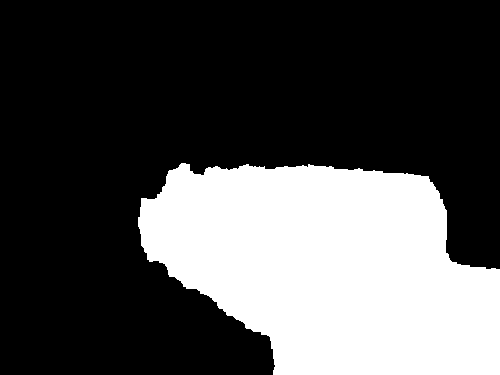} & \hspace{-0.4cm}
			\includegraphics[width=.115\textwidth]{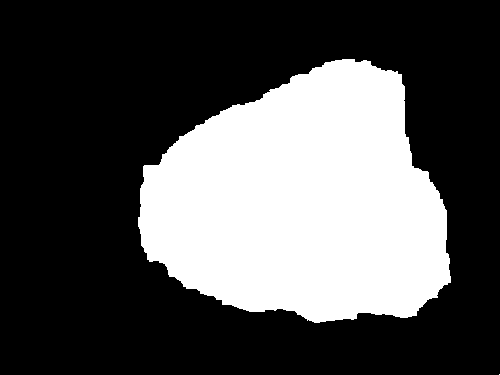}& \hspace{-0.3cm}
			\includegraphics[width=.13\textwidth]{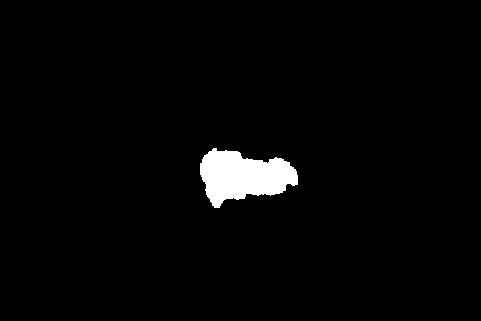} & \hspace{-0.4cm}
			\includegraphics[width=.13\textwidth]{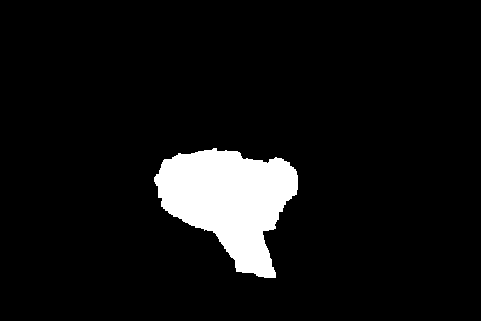} & \hspace{-0.3cm}
			\includegraphics[width=.1\textwidth]{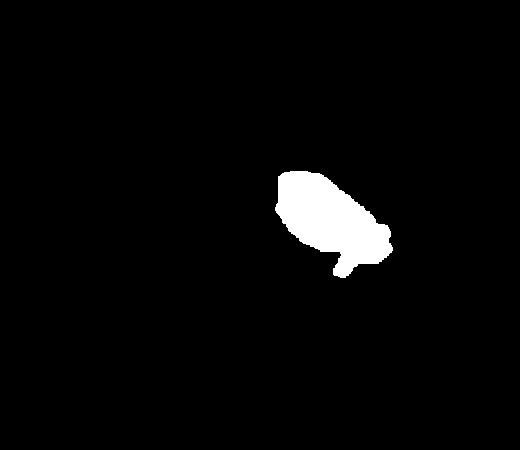} & \hspace{-0.4cm}
			\includegraphics[width=.1\textwidth]{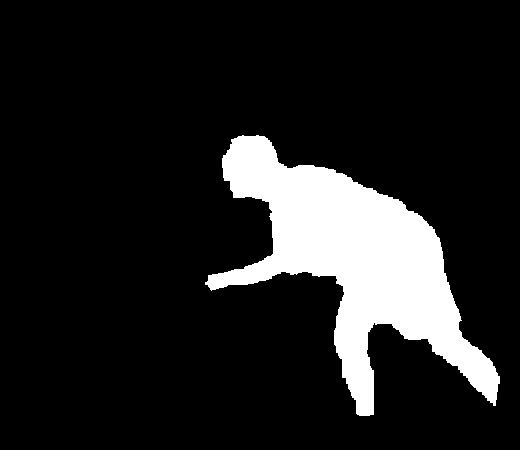}\\[-0.15in]
			{\rotatebox[origin=t]{90}{\hspace{1cm}TVPD}}&\hspace{-0.25cm}\includegraphics[width=.115\textwidth]{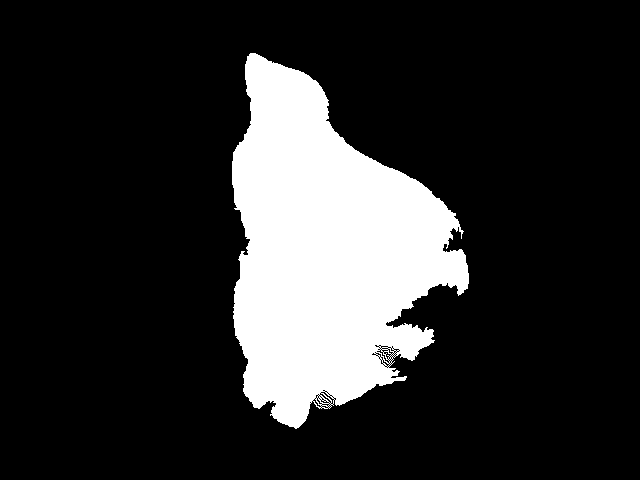} & \hspace{-0.4cm}
			\includegraphics[width=.115\textwidth]{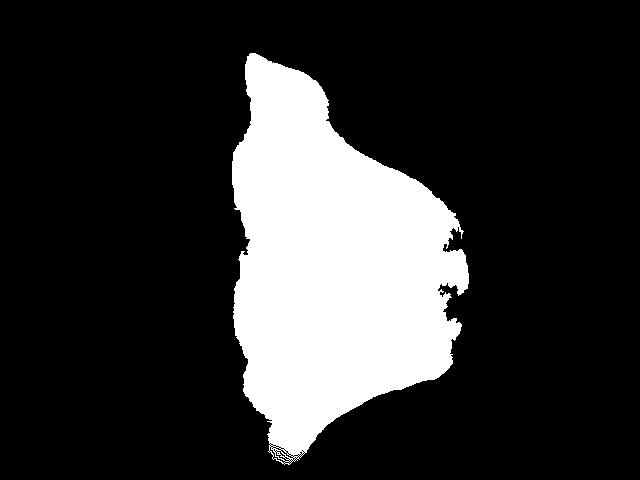} &\hspace{-0.3cm} 
			\includegraphics[width=.115\textwidth]{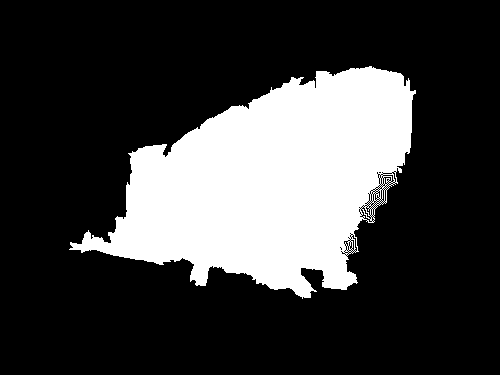} & \hspace{-0.4cm}
			\includegraphics[width=.115\textwidth]{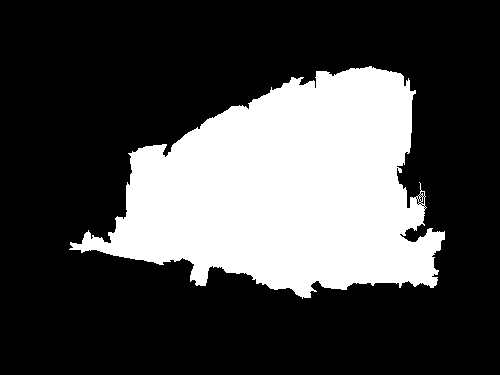}& \hspace{-0.3cm}
			\includegraphics[width=.13\textwidth]{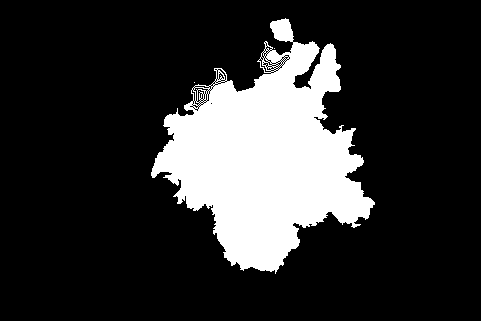} & \hspace{-0.4cm}
			\includegraphics[width=.13\textwidth]{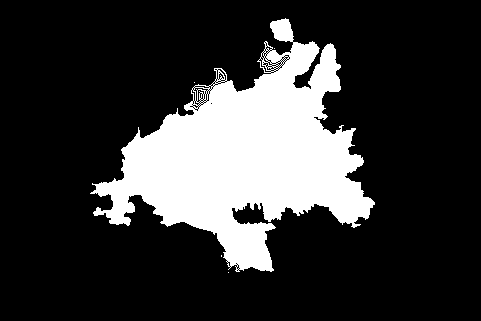} &\hspace{-0.3cm} 
			\includegraphics[width=.1\textwidth]{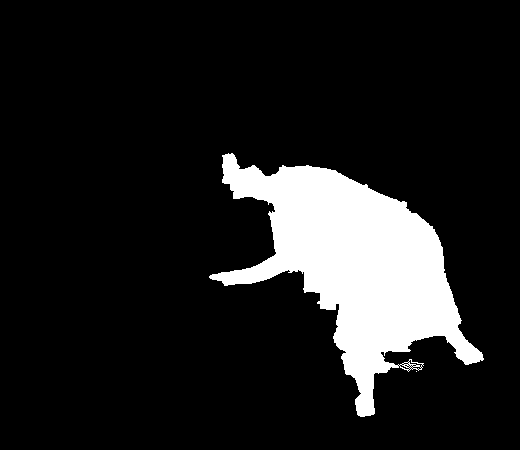} & \hspace{-0.4cm}
			\includegraphics[width=.1\textwidth]{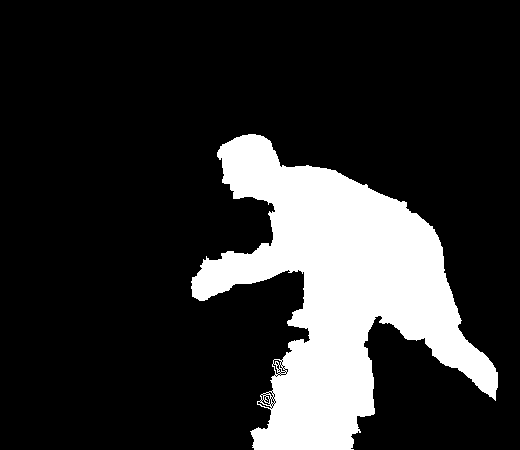}\\[-0.24in]
			{\rotatebox[origin=t]{90}{\hspace{1cm}TVAD}}&\hspace{-0.25cm}\includegraphics[width=.115\textwidth]{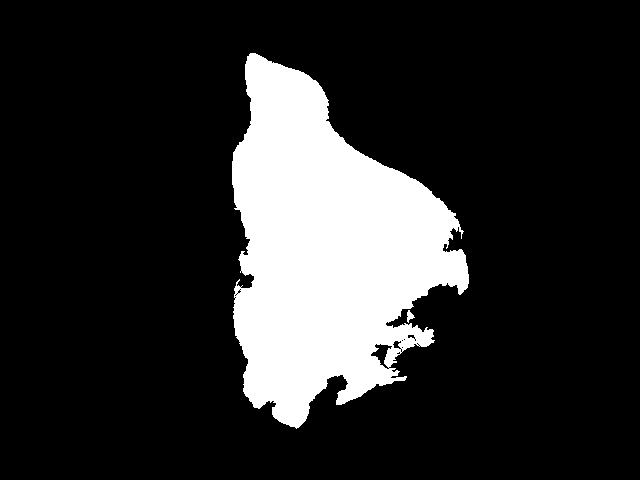} &\hspace{-0.4cm} 
			\includegraphics[width=.115\textwidth]{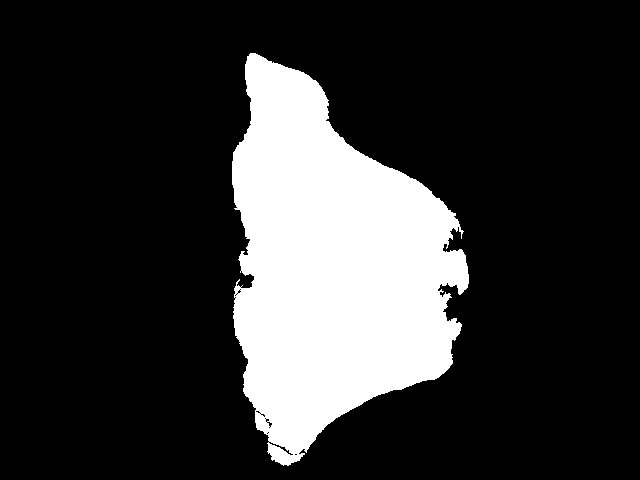} & \hspace{-0.3cm}
			\includegraphics[width=.115\textwidth]{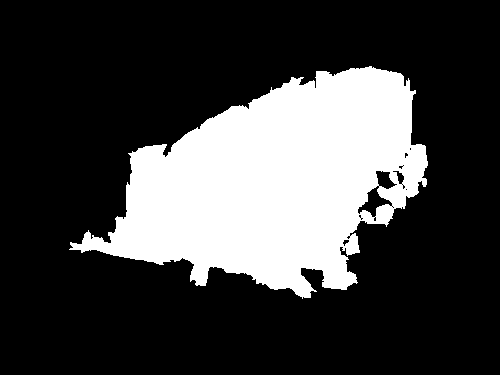} & \hspace{-0.4cm}
			\includegraphics[width=.115\textwidth]{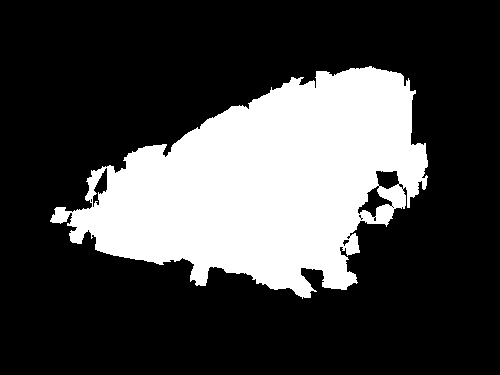}& \hspace{-0.3cm}
			\includegraphics[width=.13\textwidth]{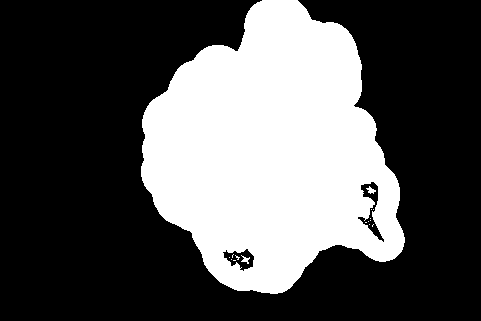} &\hspace{-0.4cm} 
			\includegraphics[width=.13\textwidth]{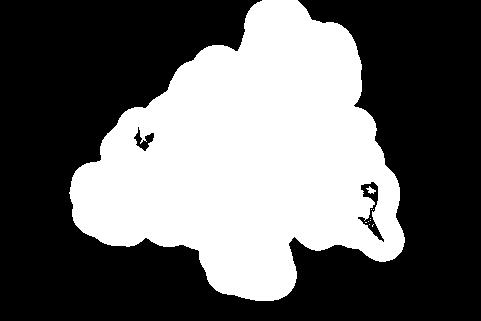} & \hspace{-0.3cm}
			\includegraphics[width=.1\textwidth]{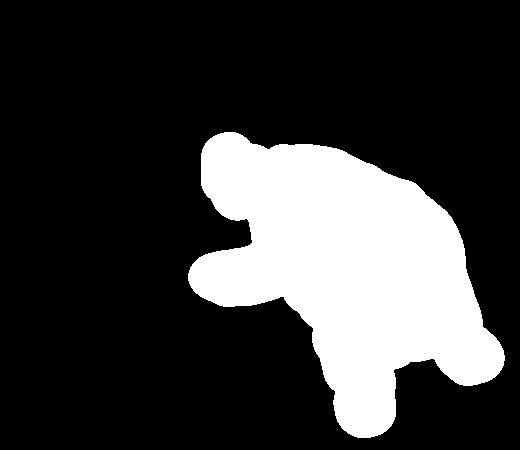} & \hspace{-0.4cm}
			\includegraphics[width=.1\textwidth]{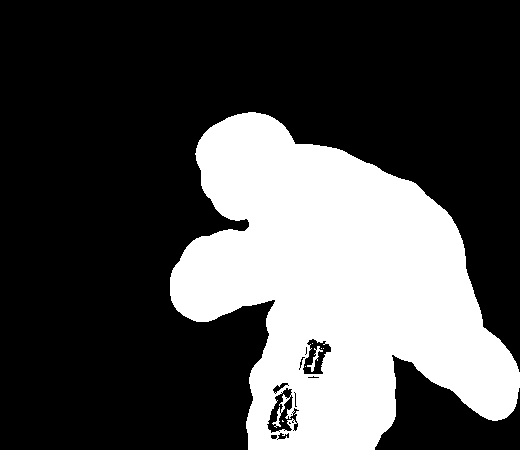}\\
		\end{tabular}
		\vspace{-0.6cm}
		\caption{Segmentation results by different methods for some images from the VGG  interactive image segmentation dataset with original annotations and modified annotations, respectively.}
		\label{fig2}
	\end{center}
\end{figure*}

\begin{figure}[!htbp]
	\centerline{\includegraphics[width=.5\textwidth]{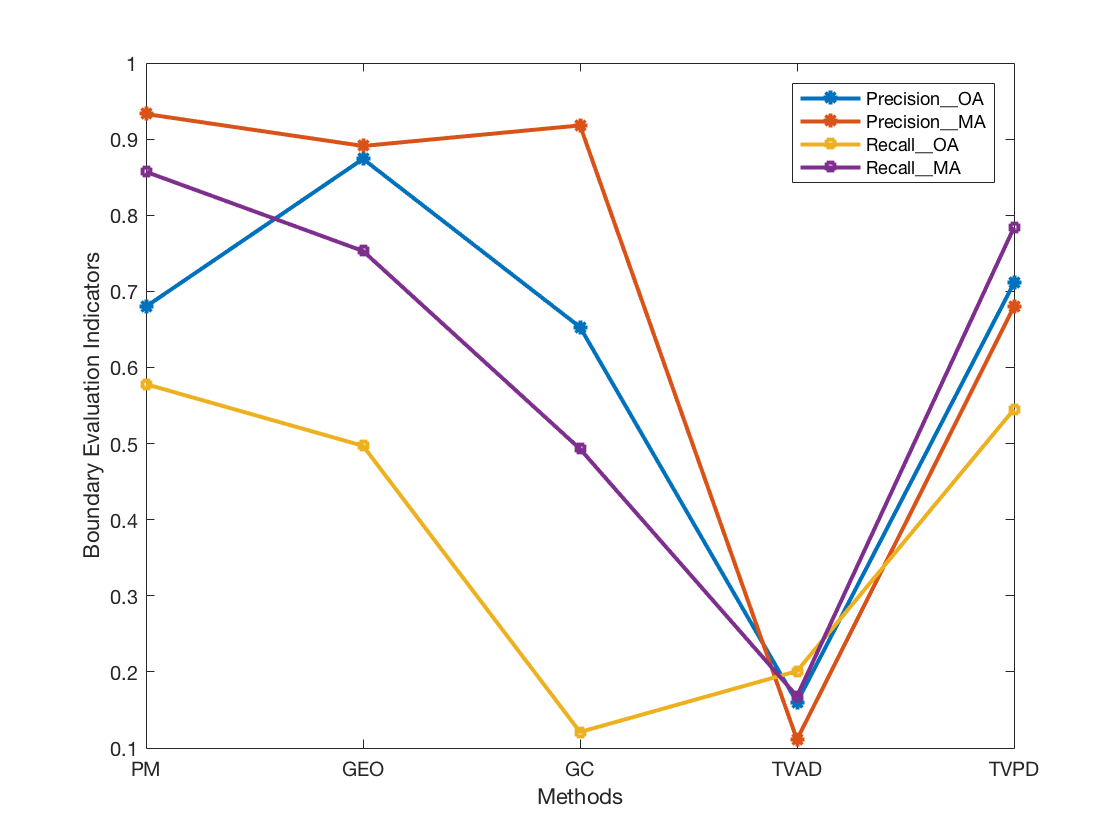}}
	\caption{Comparisons of  the average \emph{boundary precision} and \emph{boundary recall} of the segmentation results by different methods for the VGG  interactive image segmentation dataset with original annotations and modified annotations, respectively.} 
	\vspace{-0.5cm}
	\label{fig:BPR}
\end{figure}

We first compare different segmentation methods using the VGG dataset with original annotations (OA).  Figure \ref{fig2} presents some examples for the visual comparisons among different methods, and Table \ref{table1} reports  the average values of \emph{IoU},  \emph{$F_2$-score} and \emph{error rate} of these methods over the whole data set. Note that the higher \emph{IoU} and  \emph{$F_2$-score} are, the better the segmentation results. 
The \emph{IoU}  of {PM} is  0.623, which is higher than all other methods whose \emph{IoU} ranges from 0.476 to 0.613. 
As to the \emph{$F_2$-score}, PM achieves 0.812 and  also shows a dominating advantage  upon other methods. The \emph{error rate} of PM is 7.91\%, which is lower than other methods.  
To demonstrate better segmentation performance, we also compare these segmentation methods using the dataset with modified annotations (MA). Due to more {\em a prior} information provided by the foreground and background seeds,  the performance of segmentation have been augmented for all methods. At the same time, PM still performs the best  among all methods, whose \emph{IoU} now  reaches 0.834, \emph{$F_2$-score} 0.930 and \emph{error rate} only $3.47\%$.   The results of average \emph{boundary precision} and \emph{boundary recall} by these methods are compared in Figure \ref{fig:BPR}, which shows that PM  almost outperforms all other compared methods on the edge-preserving ability except  \emph{boundary precision} for the dataset with original annotations.

\begin{table}[!h]
	\centering
	\begin{tabular}{|c|l|ccr|}\hline    
		Annotations & Method                   &\emph{IoU}           & \emph{$F_2$-score}   & \emph{error rate}  \\\hline\hline
		\multirow{7}{*}{OA}& \textbf{PM} & \textbf{0.623} & \textbf{0.812} & \textbf{7.91\%} \\
		&GEO                    & 0.595          & 0.764     & 9.07\%         \\
		&GC               & 0.476          & 0.692     & 10.11\%         \\
		&TVPD                   & 0.607          & 0.797     & 8.29\%         \\
		&TVAD                   & 0.613          & 0.802     & 8.17\%         \\\hline\hline
		\multirow{7}{*}{MA}&\textbf{PM} & \textbf{0.834} & \textbf{0.930} & \textbf{3.47\%} \\ 
		% &PM-SP800               & 0.754          & 0.880      & 4.787E-2 \\
		% &PM w/o FBS             & 0.800          & 0.902      & 5.109E-2 \\
		&GEO                    & 0.744          & 0.849      & 5.57\% \\
		&GC               & 0.685          & 0.812      & 6.32\% \\
		&TVPD                   & 0.730          & 0.842      & 5.54\% \\
		&TVAD                   & 0.753          & 0.856      & 5.39\% \\\hline
	\end{tabular}
	\caption{Average values of \emph{IoU},  \emph{$F_2$-score} and \emph{error rate} of the segmentation results by different methods for the VGG  interactive image segmentation dataset with original annotations and modified annotations, respectively.} 
	\label{table1}
\end{table}

\subsection*{Ablation study}

We now test how the number of superpixels and the choice of regularizers affect the performance of the proposed method.

\subsubsection{Effect of the number of superpixels}
\begin{figure}[!h]
	\begin{center}
		\hspace*{-.25cm}
		\begin{tabular}{ccc}
			RGB & \hspace{-0.4cm}  PM-SP800 & \hspace{-0.4cm}  PM-SP1600\\
			\includegraphics[width=.15\textwidth]{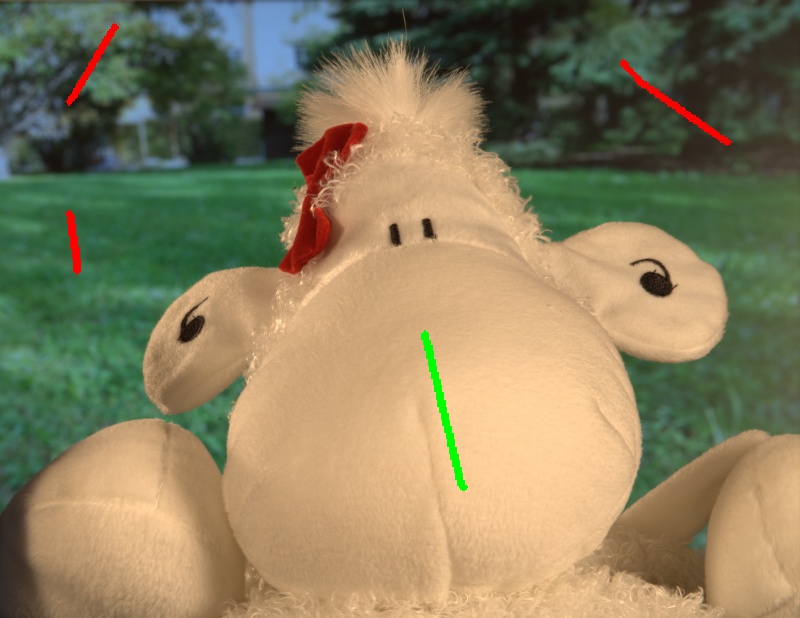} &\hspace{-0.4cm} 
			\includegraphics[width=.15\textwidth]{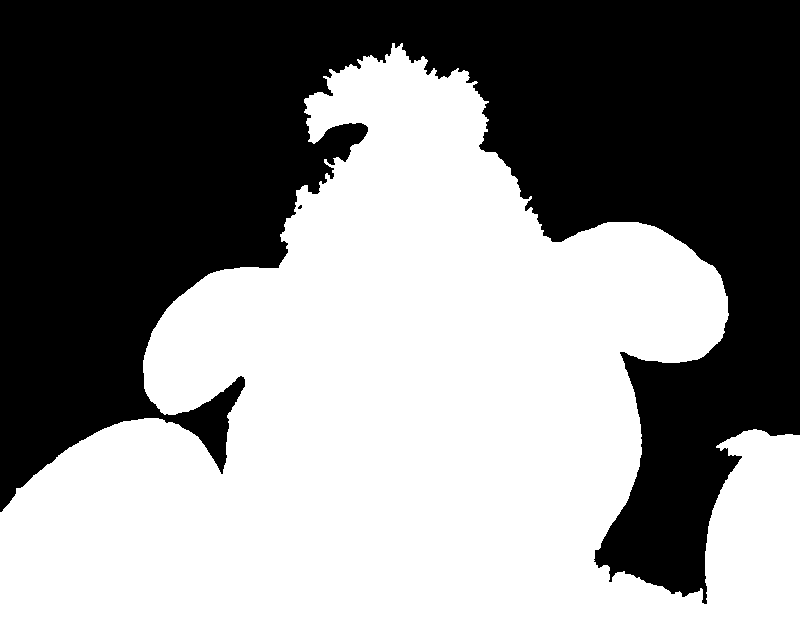} &\hspace{-0.4cm} 
			\includegraphics[width=.15\textwidth]{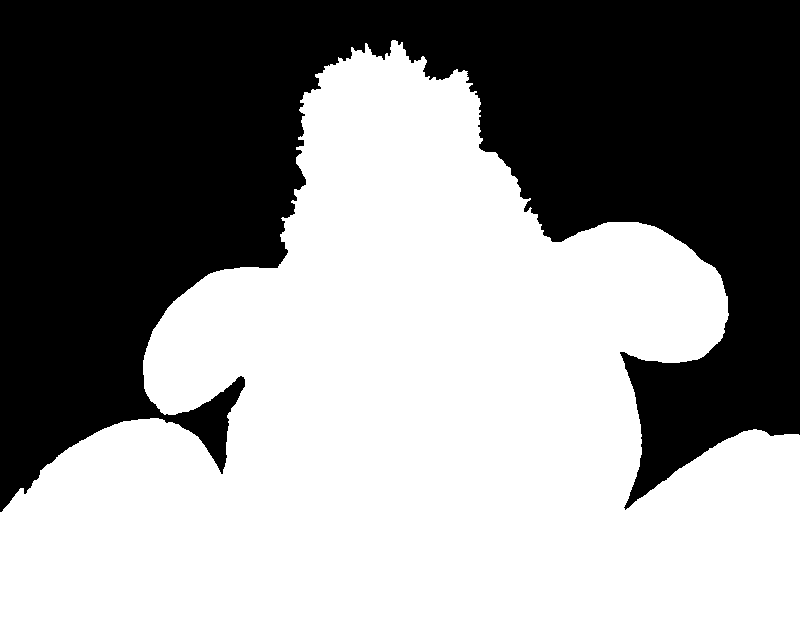}\\
			\includegraphics[width=.15\textwidth]{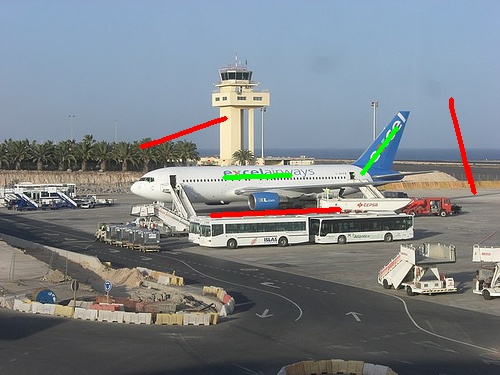} & \hspace{-0.4cm}
			\includegraphics[width=.15\textwidth]{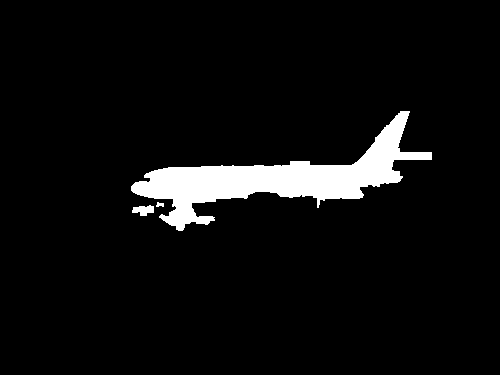} & \hspace{-0.4cm}
			\includegraphics[width=.15\textwidth]{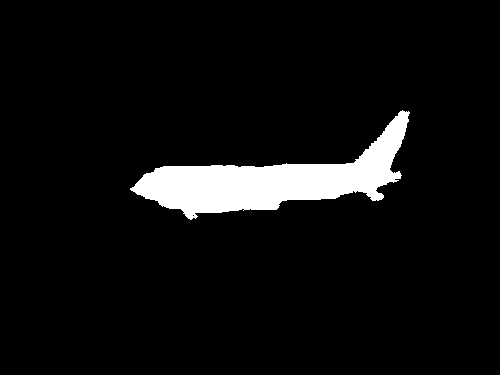}\\
		\end{tabular}
		\caption{Two segmentation examples using 800 and 1600 superpixels in  the proposed method, respectively.}
		\label{fig4}
	\end{center}
\end{figure}

\begin{figure}[!h]
	\centerline{\includegraphics[width=.5\textwidth]{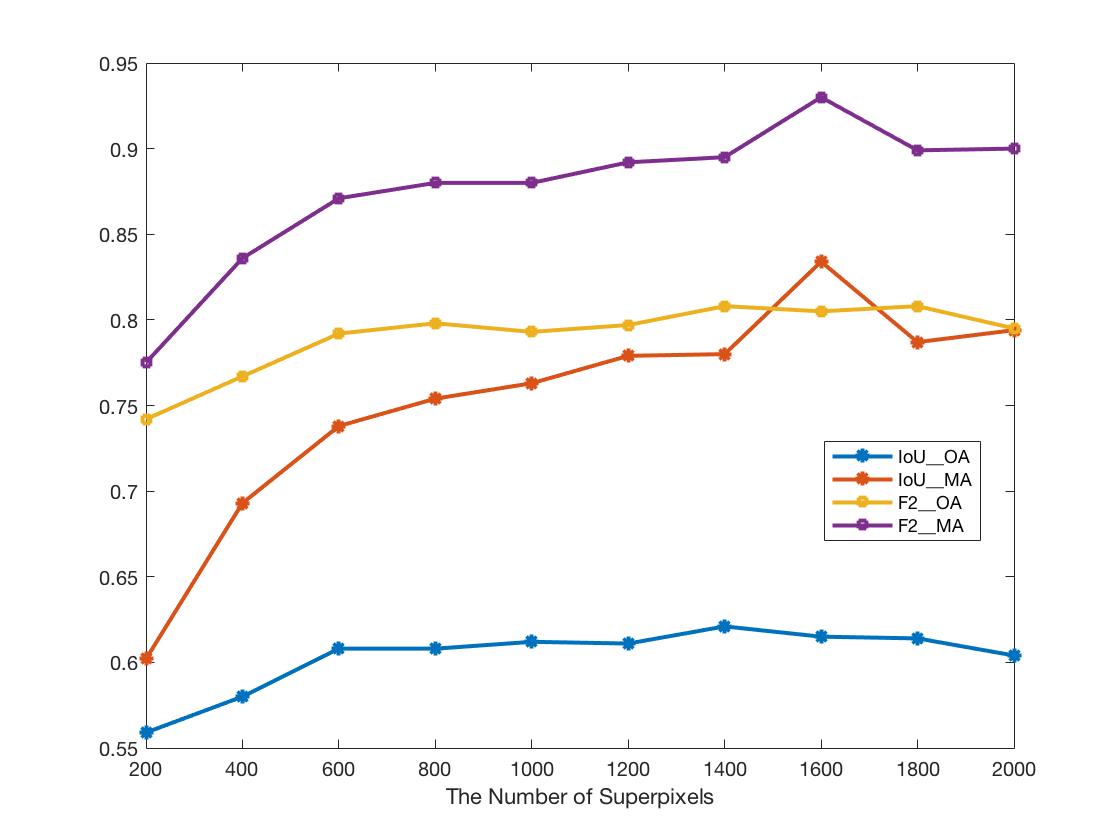}}
	\caption{Effect of the number of  superpixels on the average {\em IoU} and {\em $F_2$-score} in the proposed method for the VGG  interactive image segmentation dataset with original annotations and modified annotations, respectively.} % \textcolor{blue}{!!!Redraw this figure by changing mIoU to IoU!!!}} 
	\label{fig5}
\end{figure}

We now set different numbers of superpixels for the proposed method and compare the segmentation results. As shown in Figure \ref{fig4}, there are some vacancies around the object edges in the  segmentations when the number of superpixels is set be 800. When the number of superpixels raises to 1600, all vacancies  have been filled up and the segmentation results become much more monolithic. 
The choice of superpixels' number should be compromised, because less superpixels can significantly decrease the computational cost of geodesic distances but make the segmentation performance be poorer, and more superpixels can lead to better segmentations but require more computation times. To this end, we investigate  the effect of superpixels' number on the segmentation performance quantitatively. Figure \ref{fig5} shows that the average values of {\em IoU} and {\em $F_2$-score}  increases slowly when the number of superpixels grows, and when the number of superpixels is more than 1600, the segmentation performances almost have no improvements. Therefore, we suggest to set the number of superpixels as around 1600 in order to ensure the segmentation performance and computational efficiency simultaneously.

\subsubsection{Effect of different regularizers}
\begin{figure}[!h]
	\begin{center}
		\hspace*{-.25cm}
		\begin{tabular}{ccc}
			RGB &  \hspace{-0.4cm}  PM w/o FBS &\hspace{-0.4cm}   PM\\
			\includegraphics[width=.155\textwidth]{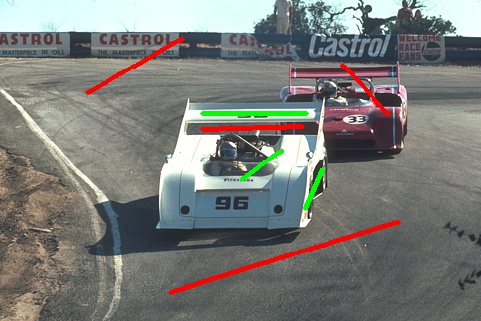} & \hspace{-0.4cm}
			\includegraphics[width=.155\textwidth]{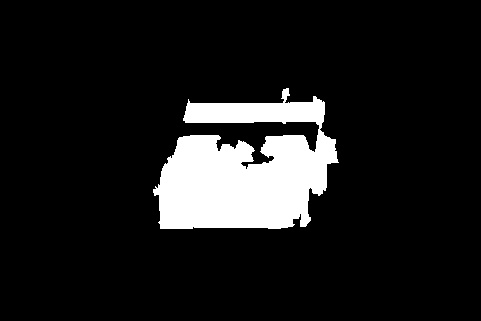} &  \hspace{-0.4cm}
			\includegraphics[width=.155\textwidth]{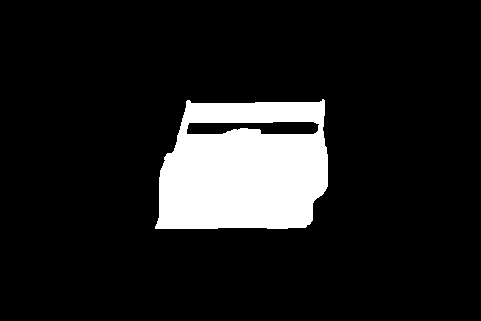}\\
			\includegraphics[width=.155\textwidth]{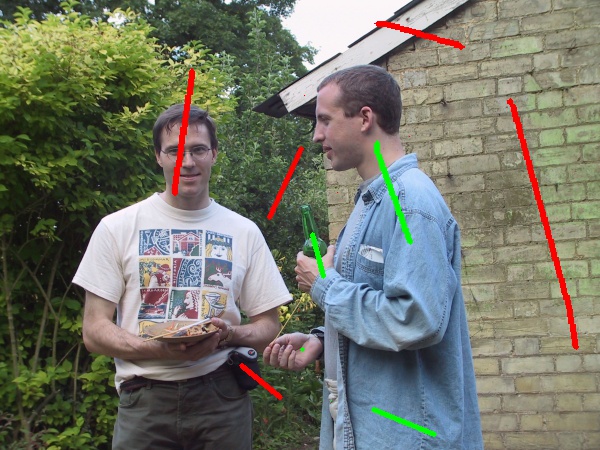} &  \hspace{-0.4cm}
			\includegraphics[width=.155\textwidth]{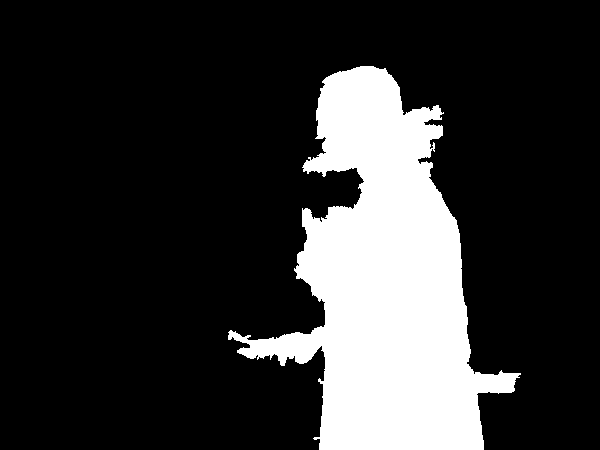} &  \hspace{-0.4cm}
			\includegraphics[width=.155\textwidth]{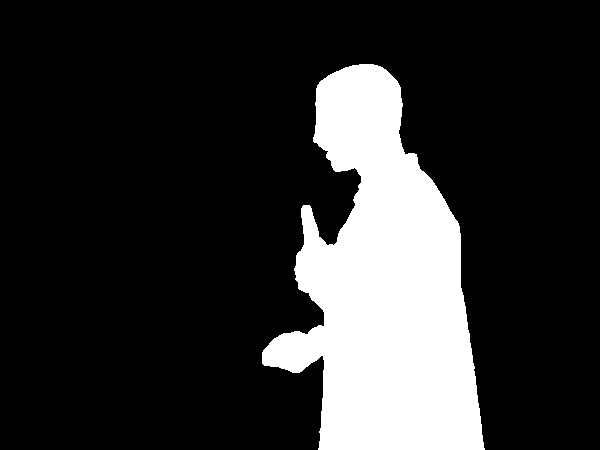}\\
		\end{tabular}
		\caption{Effect of the FBS module on the segmentation performance in the proposed method.}
		\label{fig6}
	\end{center}
\end{figure}

In the proposed method, we use the bilateral affinity  as a regularizer and efficiently compute it by using the FBS module. We first test the segmentation performance in two cases, one is combined with the FBS module, and the other is without the FBS module. 
Figure \ref{fig6} shows the segmentation results of these two cases  with 1600 superpixels for two example images. This experiment is run on the dataset with modified annotations.  We can easily see that the FBS module can greatly help to make the edge of object much more continuous and smooth.

\begin{figure}[!h]
	\begin{center}
		\hspace*{-.25cm}
		\begin{tabular}{ccccc}
			RGB &\hspace{-0.4cm}  GT &\hspace{-0.4cm}   PM &\hspace{-0.4cm}   TVPD &\hspace{-0.4cm}   TVAD\\
			\includegraphics[width=.09\textwidth]{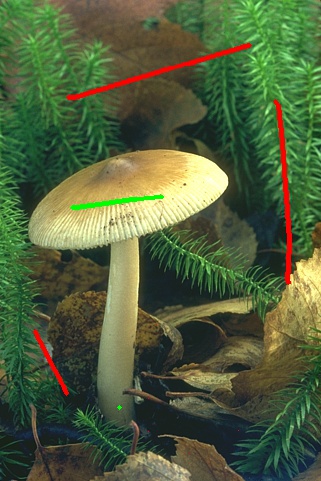}      &  \hspace{-0.4cm}
			\includegraphics[width=.09\textwidth]{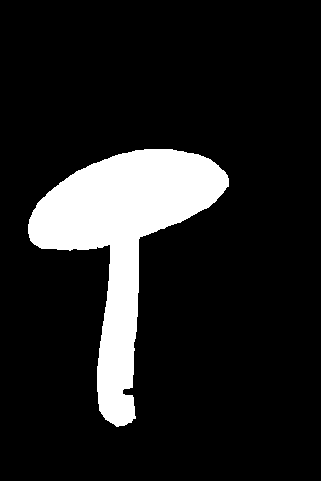}   & \hspace{-0.4cm}
			\includegraphics[width=.09\textwidth]{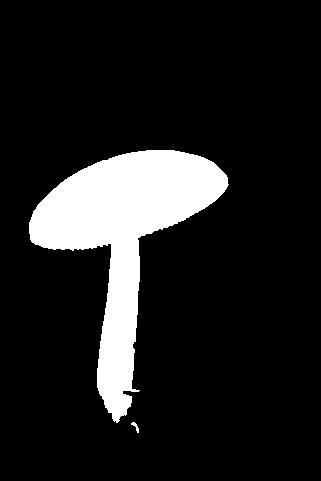}   & \hspace{-0.4cm}
			\includegraphics[width=.09\textwidth]{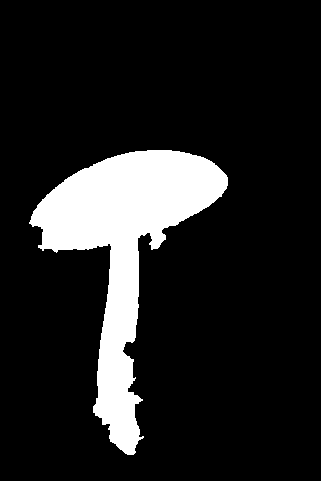} & \hspace{-0.4cm}
			\includegraphics[width=.09\textwidth]{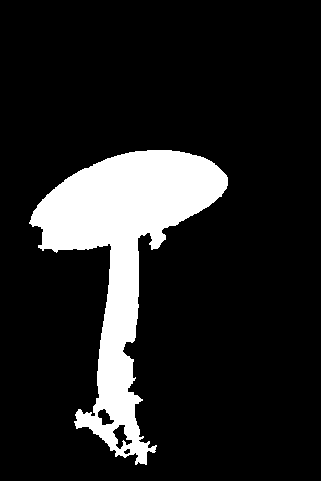}\\
			\includegraphics[width=.09\textwidth]{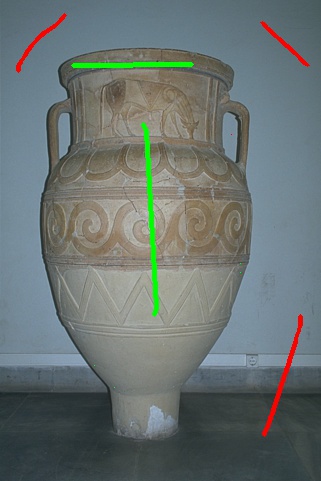}      & \hspace{-0.4cm}
			\includegraphics[width=.09\textwidth]{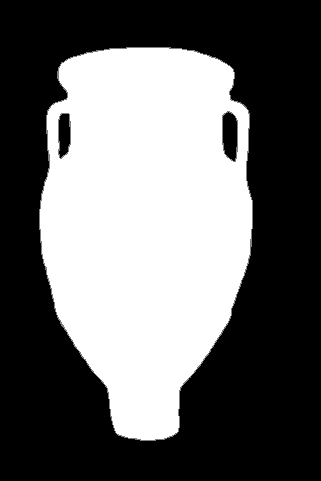}   & \hspace{-0.4cm}
			\includegraphics[width=.09\textwidth]{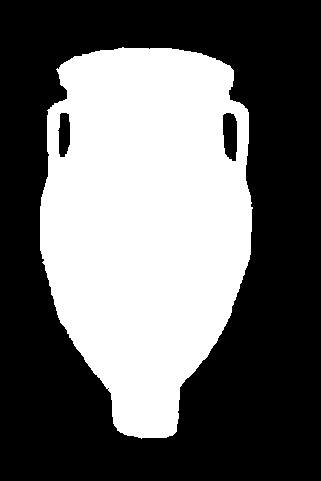}   & \hspace{-0.4cm}
			\includegraphics[width=.09\textwidth]{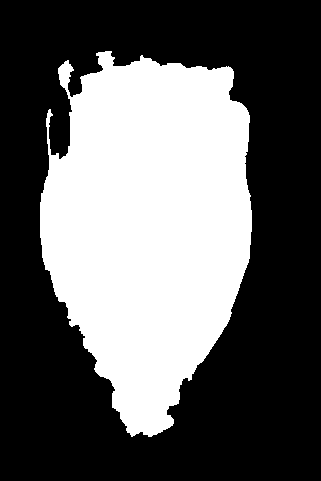} & \hspace{-0.4cm}
			\includegraphics[width=.09\textwidth]{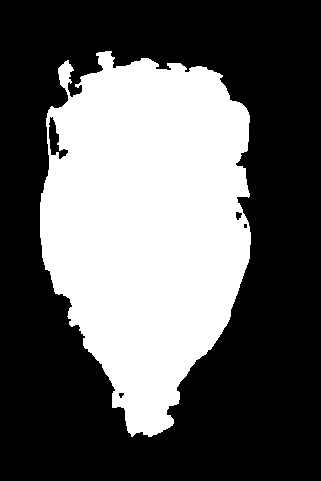}\\
			\includegraphics[width=.09\textwidth]{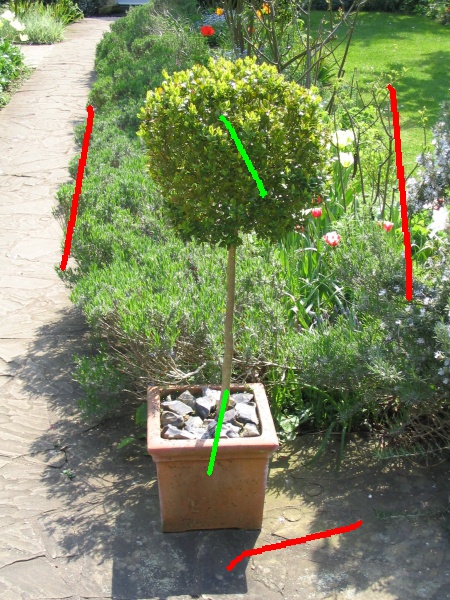}      & \hspace{-0.4cm}
			\includegraphics[width=.09\textwidth]{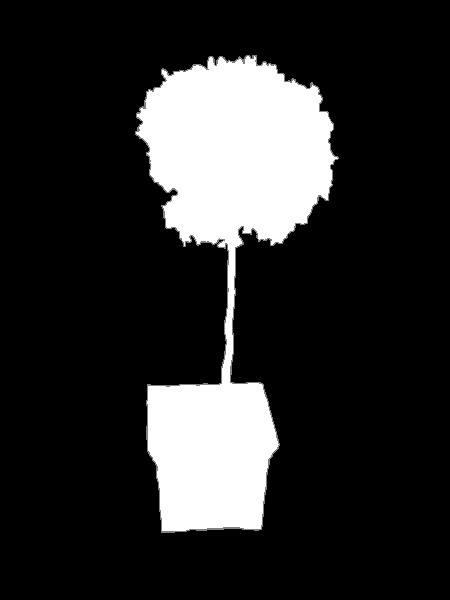}   & \hspace{-0.4cm}
			\includegraphics[width=.09\textwidth]{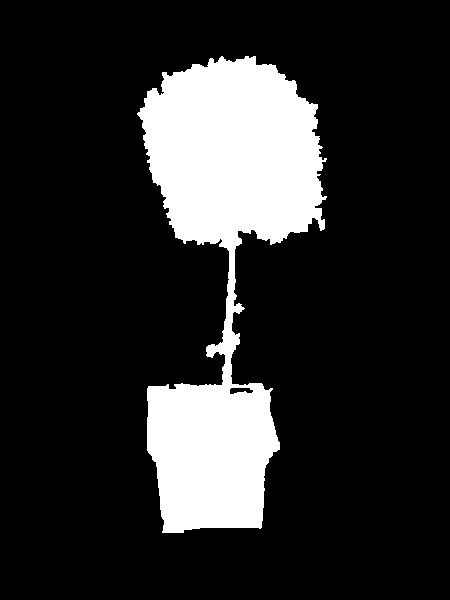}   & \hspace{-0.4cm}
			\includegraphics[width=.09\textwidth]{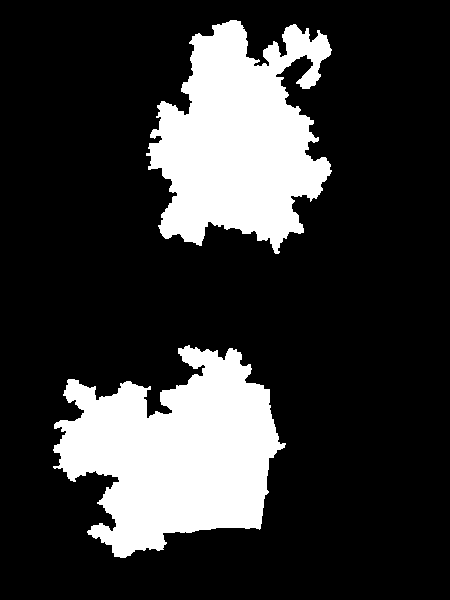} & \hspace{-0.4cm}
			\includegraphics[width=.09\textwidth]{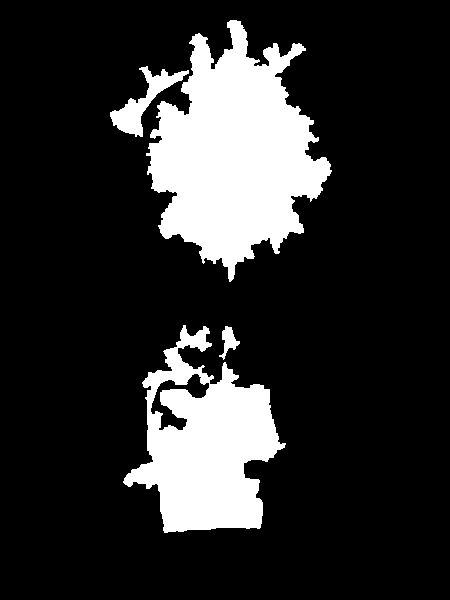}\\
			\includegraphics[width=.09\textwidth]{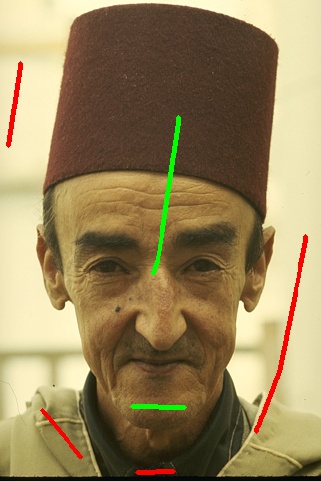}      & \hspace{-0.4cm}
			\includegraphics[width=.09\textwidth]{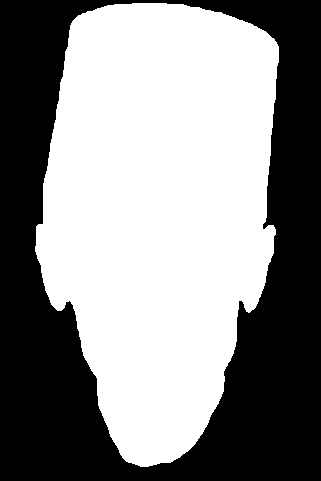}   & \hspace{-0.4cm}
			\includegraphics[width=.09\textwidth]{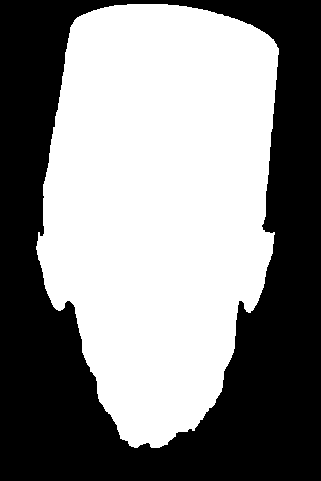}   &  \hspace{-0.4cm}
			\includegraphics[width=.09\textwidth]{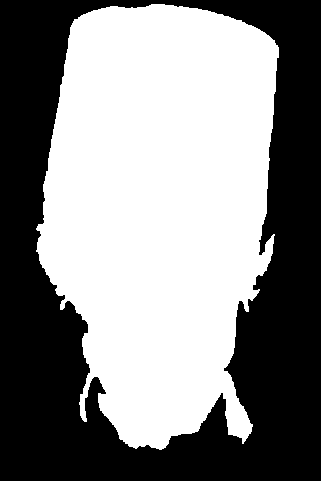} &  \hspace{-0.4cm}
			\includegraphics[width=.09\textwidth]{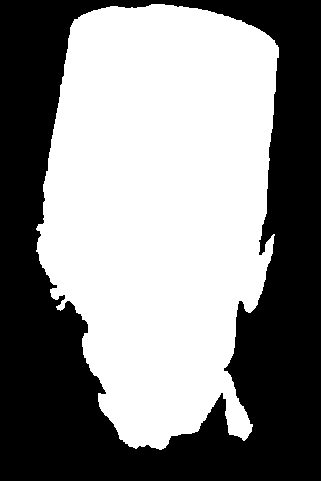}\\
		\end{tabular}
		\caption{Comparison of the performance of regularizers on edge preservation: the FBS module and the TV module.}
		\label{fig7}
	\end{center}
\end{figure}

There exist many other regularizers to preserve the edge information in the segmentation. For the comparison with FBS, we take the commonly used total variation (TV) module to show the influence of different regularizer on segmentation results. Several examples are presented in Figure \ref{fig7}, which demonstrates better performance of the FBS module over the TV module to some extent.

% We can find out that the FBS module preserves the edge better than the TV module, both the TVAD and TVPD. Also, the FBS module can help the algorithm segment the %narrow area while the two TV modules can not. And we can claim that the FBS module make a much more sense in the segmentation than the TV module in our proposed %method.

\section{Conclusion}

We present an interactive binary segmentation method based on the MRF framework, which contains the unary term and the pairwise term. 
The unary term is constructed  by  using the geodesic distances based on superpixels, which can help reduce the sensitivity to the seed placement. To relax the computational burden,  the geodesic distances  from the center of each superpixel to the foreground and background seed sets is computed instead of that from each pixel to the sets. Furthermore, we use the bilateral affinity as a regularizer to generate the pairwise term, which can denoise and well preserve the edge information. To finally solve the energy minimization problem, we take the alternative direction strategy to split it into two subproblems, which can be effectively solved by  SGD and FBS, respectively. Experimental results  on the VGG interactive image segmentation dataset  demonstrates that the proposed method is able to obtain satisfactory segmentations for a variety of images and could outperform several state-of-the-art ones according to the comparisons  in the paper. 

\newpage
\newpage
\bibliographystyle{aaai}
\bibliography{reference}

\end{document}